\documentclass{article}

\usepackage{microtype}
\usepackage{graphicx}
\usepackage{subfigure}
\usepackage{booktabs} 

\usepackage{hyperref}

\usepackage[accepted]{icml2023}

\usepackage{amsmath}
\usepackage{amssymb}
\usepackage{mathtools}
\usepackage{amsthm}

\usepackage[capitalize,noabbrev]{cleveref}

\theoremstyle{plain}
\newtheorem{theorem}{Theorem}[section]
\newtheorem{proposition}[theorem]{Proposition}

\theoremstyle{definition}
\newtheorem{definition}[theorem]{Definition}

\theoremstyle{remark}

\usepackage{amsmath,amsfonts,bm}

\newcommand{\RNum}[1]{\uppercase\expandafter{\romannumeral #1\relax}}
\newcommand{\Rnum}[1]{\lowercase\expandafter{\romannumeral #1\relax}}

\def\Secref#1{Section~\ref{#1}}
\def\Appref#1{Appendix~\ref{#1}}

\def\Figref#1{Figure~\ref{#1}}

\def\Tabref#1{Table~\ref{#1}}

\def\Secref#1{Section~\ref{#1}}

\def\eqref#1{equation~(\ref{#1})}
\def\Eqref#1{Equation~(\ref{#1})}

\def\Algref#1{Algorithm~\ref{#1}}

\def\peqref#1{(\ref{#1})}

\def \Thmref#1{Theorem~\ref{#1}}

\def\0{\bm{0}} 
\def\1{\bm{1}}

\def\vdelta{{\bm{\delta}}} 

\def\vc{{\bm{c}}}

\def\vg{{\bm{g}}}

\def\vu{{\bm{u}}}
\def\vv{{\bm{v}}}
\def\vw{{\bm{w}}}
\def\vx{{\bm{x}}}

\DeclareMathAlphabet{\mathsfit}{\encodingdefault}{\sfdefault}{m}{sl}
\SetMathAlphabet{\mathsfit}{bold}{\encodingdefault}{\sfdefault}{bx}{n}

\usepackage[algo2e]{algorithm2e}

\SetCommentSty{mycommfont}
\usepackage{color,soul}
\usepackage{multirow}
\usepackage{bbm}

\usepackage[textsize=tiny]{todonotes}

\icmltitlerunning{Differentially Private Sharpness-Aware Training}

\begin{document}

\twocolumn[
\icmltitle{Differentially Private Sharpness-Aware Training}



\icmlsetsymbol{equal}{*}

\begin{icmlauthorlist}
\icmlauthor{Jinseong Park}{ie}
\icmlauthor{Hoki Kim}{ie,er}
\icmlauthor{Yujin Choi}{ie}
\icmlauthor{Jaewook Lee}{ie}

\end{icmlauthorlist}

\icmlaffiliation{ie}{Department of Industrial Engineering, Seoul National University, 1 Gwanak-ro,
Gwanak-gu, Seoul 08826, South Korea.}
\icmlaffiliation{er}{Institute of Engineering Research, Seoul National University, 1 Gwanak-ro, Gwanak-gu, Seoul 08826, South Korea}

\icmlcorrespondingauthor{Jaewook Lee}{jaewook@snu.ac.kr}

\icmlkeywords{Machine Learning, ICML}

\vskip 0.3in
]



\printAffiliationsAndNotice{}  

\begin{abstract}
Training deep learning models with differential privacy (DP) results in a degradation of performance. 
The training dynamics of models with DP show a significant difference from standard training, whereas understanding the geometric properties of private learning remains largely unexplored.
In this paper, we investigate sharpness, a key factor in achieving better generalization, in private learning. We show that flat minima can help reduce the negative effects of per-example gradient clipping and the addition of Gaussian noise. 
We then verify the effectiveness of Sharpness-Aware Minimization (SAM) for seeking flat minima in private learning. However, we also discover that SAM is detrimental to the privacy budget and computational time due to its two-step optimization.
Thus, we propose a new sharpness-aware training method that mitigates the privacy-optimization trade-off. Our experimental results demonstrate that the proposed method improves the performance of deep learning models with DP from both scratch and fine-tuning. Code is available at \url{https://github.com/jinseongP/DPSAT}.

\end{abstract}

\section{Introduction}
\label{introduction}
%

\begin{figure}[!t]
\centering     
    \includegraphics[width=65mm]{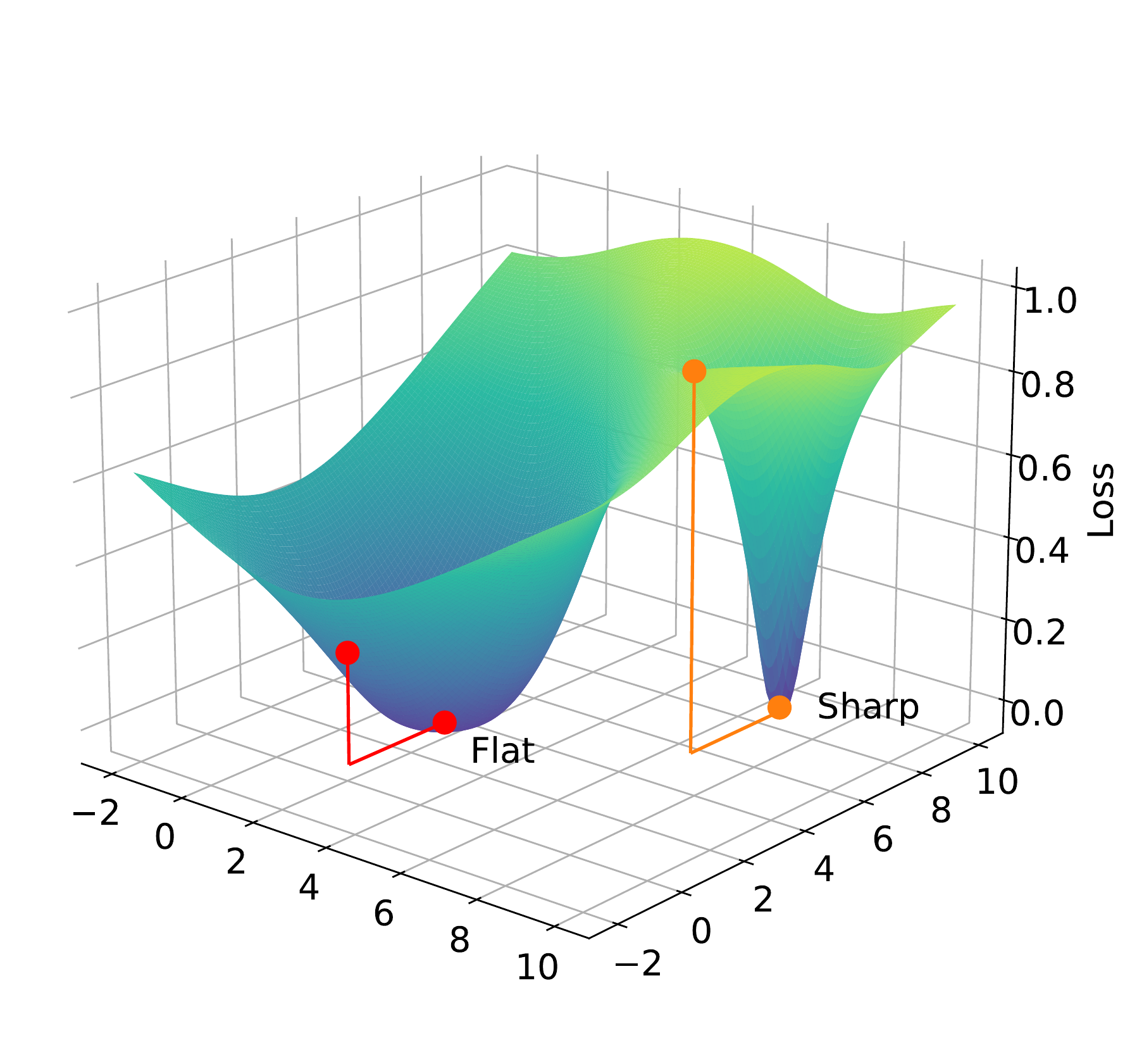}
    \caption{Illustration of flat and sharp minima. The flat minimum is robust to the sharp one given the same size of perturbation in DP training.}
    \label{fig:loss_landscape}
\end{figure}

Deep learning models are known to have a risk of privacy leakage \cite{zhu2019deep}. To protect the training data from potential data exposure, differential privacy (DP) \cite{dwork2006differential} provides a mathematical guarantee against adversaries.
Nevertheless, training deep learning models with differential privacy (DP training) can result in a degradation of prediction performance compared to models without differential privacy (non-DP training)  \cite{dwork2014algorithmic,abadi2016deep,park2023efficient}.

DP-SGD \cite{abadi2016deep} is the most popular algorithm for ensuring privacy in deep learning. The primary factors of accuracy drop in DP-SGD are known as \textbf{per-example gradient clipping} and \textbf{the addition of Gaussian noise}. 
To mitigate these effects and improve the performance of DP-SGD, many algorithmic solutions concentrate on finding the proper settings for private learning, i.e., different architectures \cite{tramer2021differentially, cheng2022dpnas}, loss functions \cite{shamsabadi2023losing}, activation functions \cite{papernot2021tempered}, and clipping functions \cite{andrew2021differentially, bu2021convergence}. 
However, achieving the ideal performance under private learning still remains an open question.



In this paper, we aim to answer \textit{``how can we find a better optimum in DP training?"} Recently, in deep learning society, it is well known that finding flat minima is a key factor for improving generalization performance \cite{keskar2017large,foret2020sharpness}.
\Figref{fig:loss_landscape} illustrates the importance of flat minimum for DP training. 
It shows that flat minimum exhibits robustness to the random perturbations, a fundamental idea of protecting data in DP methods.
Furthermore, we show that the flatness of the loss landscape can reduce the negative influences of clipping and noise addition.


Subsequently, we explore the effectiveness of optimization strategies for finding flat minima in private learning, particularly Sharpness-Aware Minimization (SAM) \cite{foret2020sharpness}. SAM, the state-of-the-art optimization method in various domains \cite{bahri2021sharpness,qu2022generalized}, efficiently finds flat minima by solving the min-max objective in two steps. 
Despite this, we point out that the two-step optimization of SAM may be detrimental to the privacy budget and computational time.
Based on these observations, we propose a new sharpness-aware training without additional privacy and computational overheads, which successfully mitigates the privacy-optimization trade-off.






Our main contributions are summarized as follows:
\begin{itemize}
    \item We demonstrate that finding flat minima can reduce the detrimental effects of clipping and noise addition in private learning. 
    \item We show the effectiveness of SAM to seek flat minima and present its drawbacks in private learning. To the best of our knowledge, this is the first attempt to study sharpness-aware training in private learning.
    
    \item  We propose \textit{Differentially Private Sharpness-Aware Training (DP-SAT)} which makes use of sharpness-aware training without additional privacy costs, achieving both generalization and time efficiency.
\end{itemize}



This paper is structured as follows: \Secref{sec:related_work} introduces  related works on DP and flat minima. \Secref{sec:dp_training} investigates how flatness helps private learning. \Secref{sec:dpsam} adapts SAM to DP and evaluates the problems caused by the two-step updates of SAM.
\Secref{sec:dpsat} introduces DP-SAT, a new while sharpness-aware training method for DP. 
\Secref{sec:exp} empirically demonstrates the effectiveness of DP-SAT across various datasets and tasks.
\Secref{sec:limit} demonstrates limitations and future works, and \Secref{sec:conclusion} concludes the paper.



 
\section{Background and Related Work}
\label{sec:related_work}

\subsection{Differentially Private Deep Learning}
Differential privacy (DP) \cite{dwork2014algorithmic} provides a formal mathematical framework to guarantee the privacy of training data. It is defined as follows:
\begin{definition}\label{def:dp}
    (Differential privacy) A randomized mechanism $\mathcal{M}:\mathcal{D}\rightarrow\mathcal{R}$ with domain $\mathcal{D}$ and range $\mathcal{R}$ satisfies $(\varepsilon,\delta)$-DP, if for two adjacent inputs $d, d'\in \mathcal{D}$ and for any set of possible outputs $\mathcal{S} \subseteq \mathcal{R}$ it holds that  
    \begin{equation}\label{eq:dp}
        Pr[\mathcal{M}(d) \in \mathcal{S}] \leq e^\varepsilon Pr[\mathcal{M}(d') \in \mathcal{S}]+\delta.
    \end{equation}
\end{definition}
The parameter $\varepsilon\geq0$ represents the privacy budget with the broken probability $\delta\geq0$. A smaller value of $\varepsilon$ implies a strong privacy guarantee of mechanism $\mathcal{M}$. In the context of deep learning, DP-SGD \cite{abadi2016deep} calculates the per-sample gradient $\nabla \ell_i(\vw)$, where $\ell_i$ is the per-data loss function of the individual data sample $\vx_i$ and $\vw$ is the model parameter. After that, it clips each per-sample gradient to a fixed $L_2$-norm and then adds Gaussian noise to the average of clipped gradients. In summary, the model weights $\vw_t$ are updated as ${\vw}_{t+1} = {\vw_t} - \eta\vg_t$ at step $t$, where the modified gradients $\vg_t$ for DP-SGD is calculated within mini-batch $I_t$ as follows:
\begin{align}
    \bar\vg_t &=\frac{1}{|I_t|}\sum_{i\in I_t}  \text{clip}(\nabla \ell_i(\vw_t),C), \label{eq:DP-SGD-nonoise} \\
    \vg_t &= \bar \vg_t+\mathcal{N}(\mathbf{0},C^2\sigma^2\mathbf{I}),
    \label{eq:DP-SGD} 
\end{align}
where \text{clip}$(\vu,C)$ projects $\vu$ to the $L_2$-ball with radius $C$ and vector norm $\|\cdot\|$ indicates the $L_2$-norm $\|\cdot\|_2$.
The noise level $\sigma$ is determined by the privacy budget $(\varepsilon,\delta)$ as follows:
\begin{proposition}
(\citet{abadi2016deep}). There exist constant $c_1$ and $c_2$ so that given total steps $T$ and sampling probability $q$, for any $\varepsilon<c_1 q^2T$, DP-SGD (\ref{eq:DP-SGD}) guarantee ($\varepsilon,\delta$)-DP, for any $\delta>0$ if we choose 
\begin{equation}
    \sigma \geq c_2 \frac{q\sqrt{T\log(1/\delta})}{\varepsilon}.
    \label{eq:DP-SGD_sigma_proposition}
\end{equation}
\label{prop:dpsgd_noise}
\end{proposition}
As gradient clipping limits the sensitivity of average gradients, DP-SGD can impede the model from updating towards the dominant gradient direction \cite{papernot2021tempered}. Moreover, the addition of noise to guarantee the privacy of training data can interrupt the convergence of the model weights to the optimum \cite{yu2021do}.
Note that using larger clipping values increases alignment with the original gradients, but also increases the variance at the same time. The additional definitions and properties of DP are summarized in \Appref{app:dp_properties}.

\subsection{Flat Minima and Sharpness-Aware Minimization}



\paragraph{Sharp and flat minima}
Understanding the geometric properties of the loss landscape is a central topic for optimization in deep learning. Generally, the Hessian matrix of loss function $\mathbf{H}=\mathbf{H}_\vw :=\nabla^2 \ell(\vw)$ and its \textit{sharpness}, which is defined as its  spectral norm $\|\mathbf{H}\|_2$  (or its top eigenvalue $\lambda_{max}$), can explain the training dynamics of gradient descent \cite{cohen2021gradient}. In other words, the loss landscape in the vicinity of a flat minimum, which has small eigenvalues of the Hessian, exhibits slow variation within a neighborhood of $\vw$. Conversely, near a sharp minimum with large eigenvalues, the loss function is vulnerable to small noises \cite{li2018visualizing,keskar2017large,dinh2017sharp}, even adversarial perturbations \cite{wu2020adversarial,lee2021towards, kim2023generating}.
To find flatter minima, various optimization techniques have been proposed, such as stochastic weight averaging \cite{izmailov2018averaging} and gradient regularizer \cite{barrett2020implicit}.

\paragraph{Sharpness-aware training} 
Recently, \citet{foret2020sharpness} proposed Sharpness-Aware Minimization (SAM), which is the state-of-the-art optimization methodology in various domains. SAM minimizes the worst-case perturbations within a radius $\rho$
in the vicinity of the parameter space as follows:
\begin{equation}
    \min_\vw \max_{\|\vdelta^*\|\leq\rho} \ell(\vw+\vdelta^*). 
    \label{eq:sam_obj}
\end{equation}
 As it is difficult to identify the optimal direction $\vdelta^*$, SAM approximates the perturbation $\vdelta$ with a first-order Taylor expansion.
 Subsequently, SAM updates the model weights in two steps, described as follows:
\begin{align}
     \vw^{p}_t&=  \vw_t + \rho \vdelta_t=  \vw_t + \rho  \frac{\nabla \ell(\vw_t)}{\|\nabla \ell(\vw_t)\|},
     \label{eq:ascent}
     \\ 
     \vw_{t+1}&= \vw_t - \eta \nabla \ell(\vw^{p}_t). 
    \label{eq:descent}
\end{align}
We initially calculate the \textit{perturbed weight} $\vw^p_t$ in the ascent step \peqref{eq:ascent}, and then update the model weights towards the gradient of perturbed loss $\nabla \ell(\vw^p_t)$ in the descent step \peqref{eq:descent}. \citet{foret2020sharpness} defined the difference $\ell(\vw_t^p)-\ell(\vw_t)$ as \textit{estimated sharpness} which should be minimized to find flat minima. Note that variants of SAM have been proposed recently to boost the generalization performance \cite{kwon2021asam,zhuang2021surrogate,kim2023stability,kim2023exploring} and reduce the computation of two-step optimization \cite{du2022sharpnessaware,du2021efficient,park2022fast}.


\subsection{Loss Landscape of Private Learning}
Recent studies have investigated the unique loss landscape and training dynamics of DP-SGD in comparison to SGD. \citet{bu2021convergence} analyzed the convergence of DP training in terms of different clipping methods and noise addition. \citet{wang2021dplis} first highlighted the problem of DP-SGD being stuck in local minima due to the training instability.
The authors suggested that averaging the gradients of neighborhoods in the parameter space can achieve a smoother loss landscape and improved performance, yet this comes with a significant computational cost. Instead of averaging, \citet{shamsabadi2023losing} proposed that loss functions with smaller norm can reduce the impact of clipping and thus create a smoother loss function.

\begin{figure}[!t]
\centering     
    \includegraphics[width=80mm]{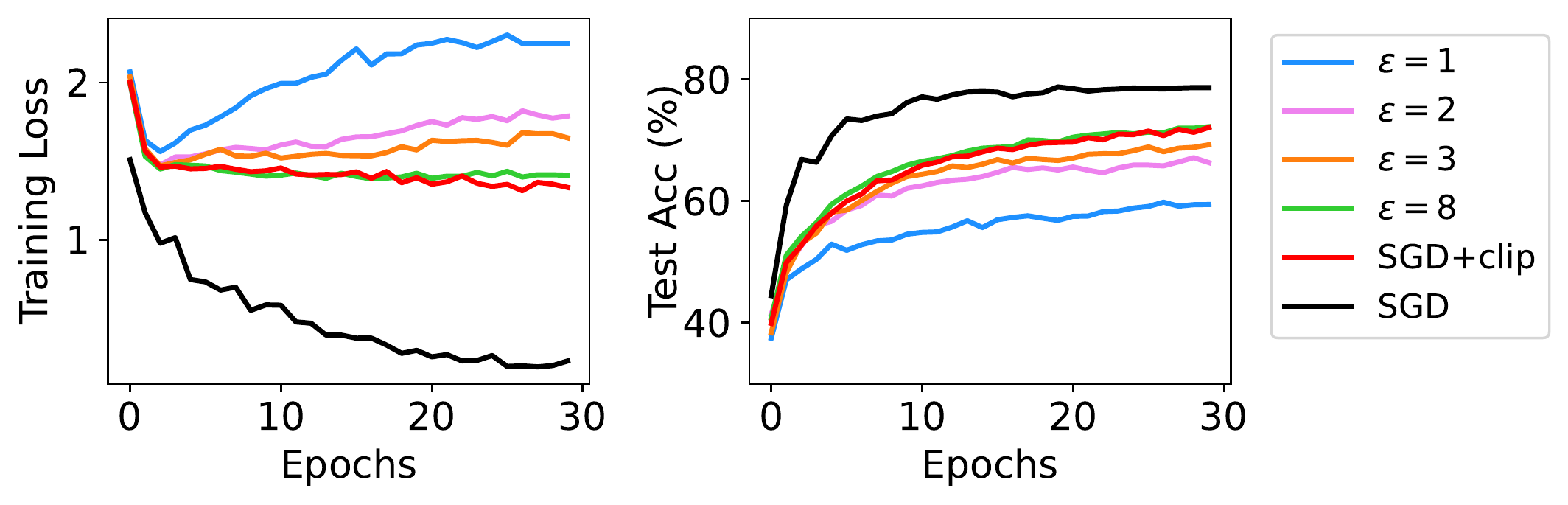}
    \caption{Training loss (left) and test accuracy (right) of DP-SGD  with $\varepsilon \in \{1,2,3,8\}$, SGD only with clipping ($C=0.1$) without noise addition (SGD+clip), and SGD on CIFAR-10.}
    \label{fig:training_loss_error}
\end{figure}
\section{Flat Minima Help Private Learning}
\label{sec:dp_training}
In this section, we first prove that achieving flatness can be beneficial to DP training. Recently, discovering the relationship between the loss function, gradient norm, and flat minima has become an important topic to analyze in deep learning optimization \cite{zhao2022penalizing, zhang2023gradient}. The interaction between these factors significantly influences generalization performance in various domains \cite{barrett2020implicit,wu2020adversarial}.
Thus, we start by investigating the learning dynamics of DP-SGD, as stated in \cite{bagdasaryan2019differential, wang2021dplis}. \Figref{fig:training_loss_error} illustrates the training loss and test accuracy of DP-SGD with various privacy budgets $\varepsilon$, SGD only with clipping (SGD+clip), and standard SGD. The training loss of DP-SGD cannot converge to zero due to the effects of clipping and noise addition. This leads to instability of training and corresponding lower performance. To mitigate this phenomenon, we investigate the vulnerability of DP training to clipping and noise addition in terms of sharpness.

\begin{figure*}[!t]
\centering     
    \subfigure[Epoch 1 (Early stage of training)]{\label{fig:gradnorm_epoch1}\includegraphics[width=55mm]{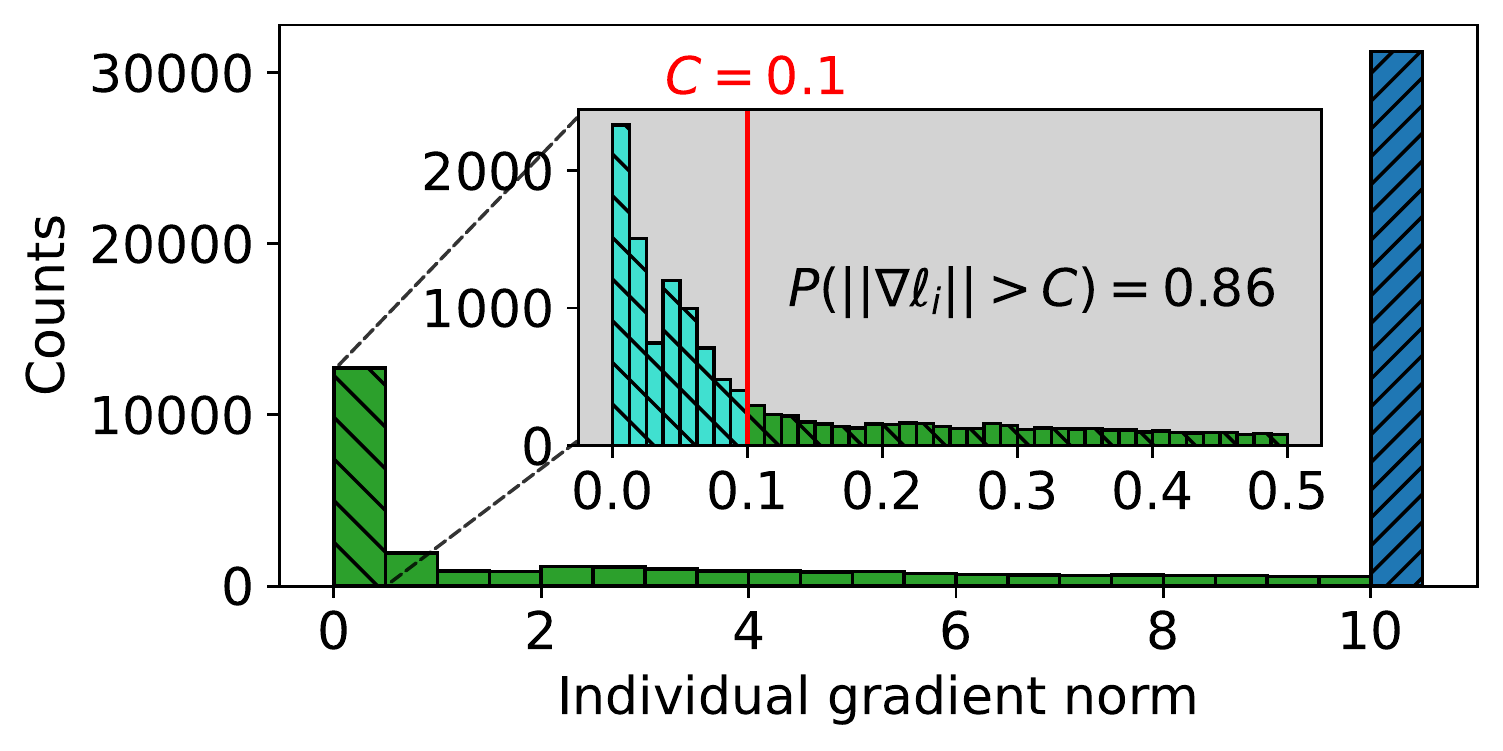}}
    \subfigure[Epoch 10]{\label{fig:gradnorm_epoch10}\includegraphics[width=56mm]{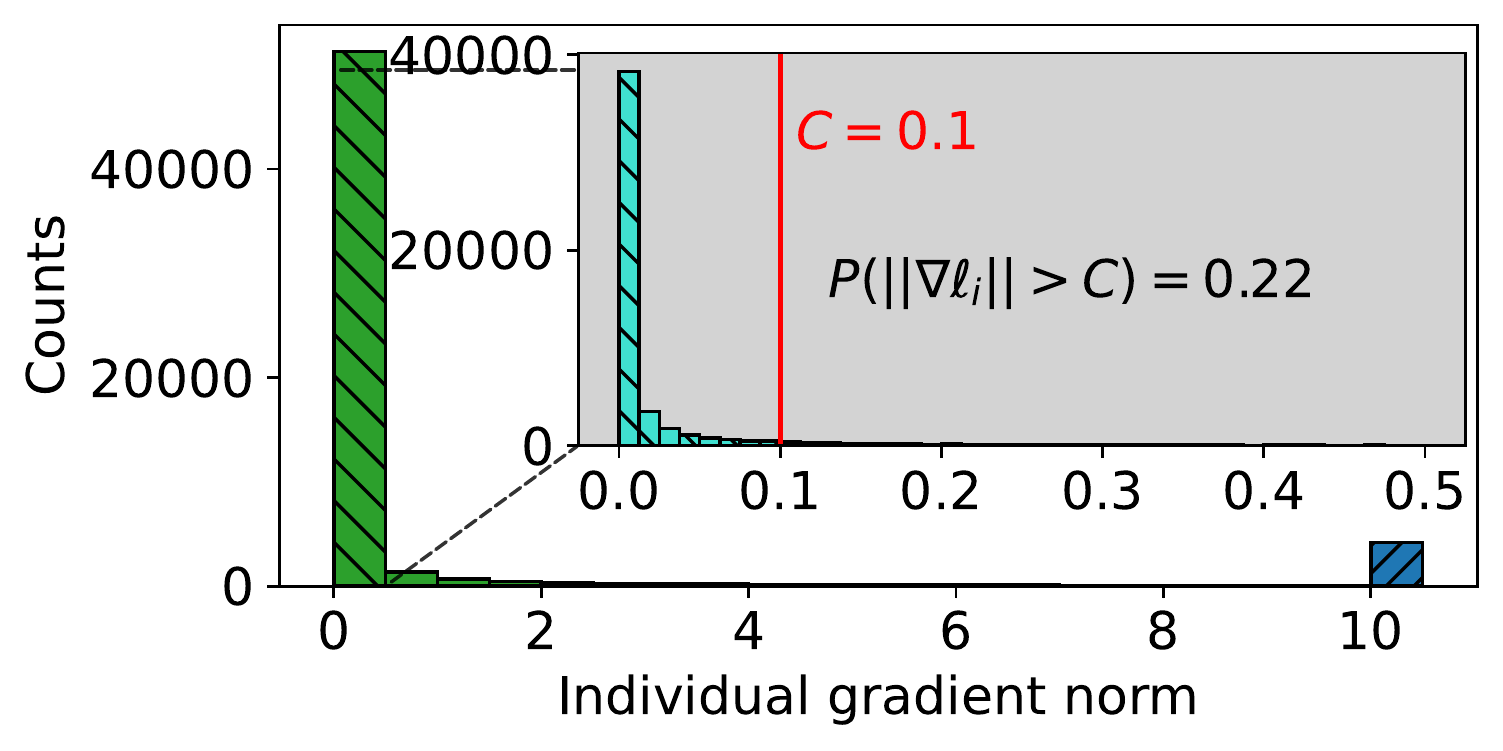}}
    \subfigure[Epoch 40 (End of training)]{\label{fig:gradnorm_epoch40}\includegraphics[width=55mm]{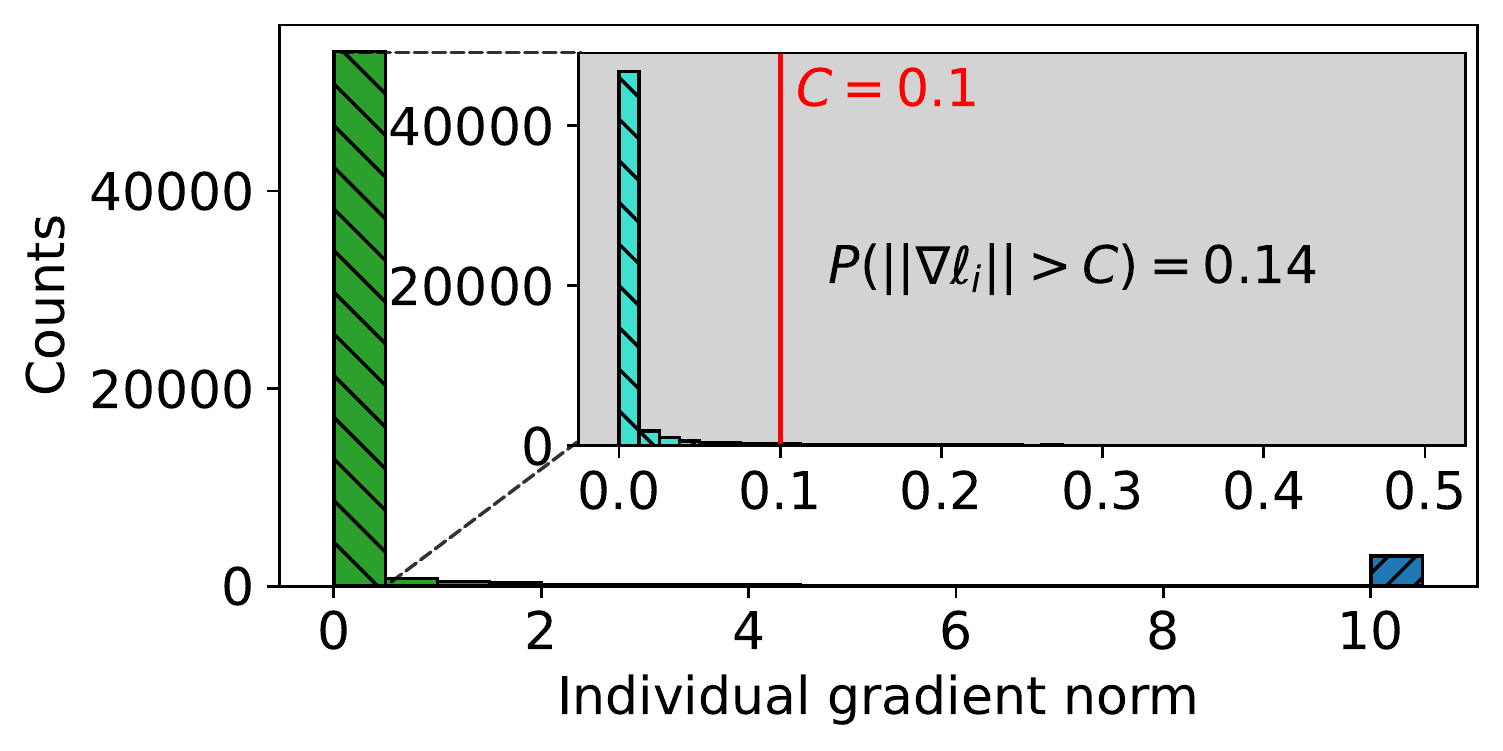}}
    \caption{Distribution of individual gradient norm $\|\nabla\ell_i\|$ for all $i$ at epoch 1 (early), 10, and 40 (end). We enlarge the values between 0 and 1 for clarifying the distribution of small gradients. The blue chart indicates the sum of all counts larger than 10. The red line indicates the gradient clipping value $C$. The proportion of being clipped is high at the early stage and decreases constantly.}
    \label{fig:grad_norm}
\end{figure*}

\subsection{Flat Minima Mitigate the Effect of Clipping} 
The difference between SGD and SGD+clip in \Figref{fig:training_loss_error} indicates that clipping itself has negative effects in training.
It means that reducing the gradient norm can improve performance by avoiding clipping \cite{papernot2021tempered}.  To this end, we argue that a flat minimum exhibits additional advantages in DP training, i.e., reducing the negative impact of clipping by bounding the gradient norm near a local optimum.


 \begin{theorem}
  \label{thm:clipping} 
 \textbf{(Flat minimum mitigates the effect of clipping)} 
The difference between gradients before and after clipping can be bounded by the sharpness as
\begin{align*}
    \|&\nabla\ell_i(\vw)-\text{clip}(\nabla \ell_i(\vw),C)\|  \\
    = & \mathbbm{1}(\|\nabla\ell_i(\vw)\|>C)\cdot(\|\nabla\ell_i(\vw)\|-C) \\
    \leq & \mathbbm{1}(\|\mathbf{H}_{\vw^*}\|_2\Delta_{\vw}>C)\cdot(\|\mathbf{H}_{\vw^*}\|_2\Delta_{\vw}-C),
\end{align*}
near a local minimum $\vw^*$, where $\|\mathbf{H}_{\vw^*}\|_2$ is the sharpness at $\vw^*$. $\Delta_{\vw} = \|\vw-\vw^*\|$ and $\mathbbm{1}$ denotes an indicator function.   
\end{theorem}
We defer the proof to \Appref{app:clipping}. As the gradient norm is upper bounded by the sharpness, a lower proportion of gradients is to be clipped within a flat minimum. 
Empirically, we measure the gradient norm of each data sample $\nabla \ell_i(\vw_t)$ trained with DP-SGD on the MNIST dataset in \Figref{fig:grad_norm}. The proportion of data samples being clipped $P(\|\nabla\ell_i(\vw)\|>C)$ is high in the early stage of training, indicating that weights are updated in a significantly different direction due to clipping. 
Even though the proportion $P(\|\nabla\ell_i(\vw)\|>C)$ diminishes as training proceeds, some individual gradients are still clipped. As these clipped gradients act as the primary direction in SGD, it is detrimental to finding an effective direction in DP-SGD.
By uncovering a flatter loss landscape, we can reduce the amount of clipped individual gradients in DP-SGD during training.

\subsection{Flat Minima Reduce the Bias of Noise}


\begin{figure}[!t]
\centering     
    \includegraphics[width=80mm]{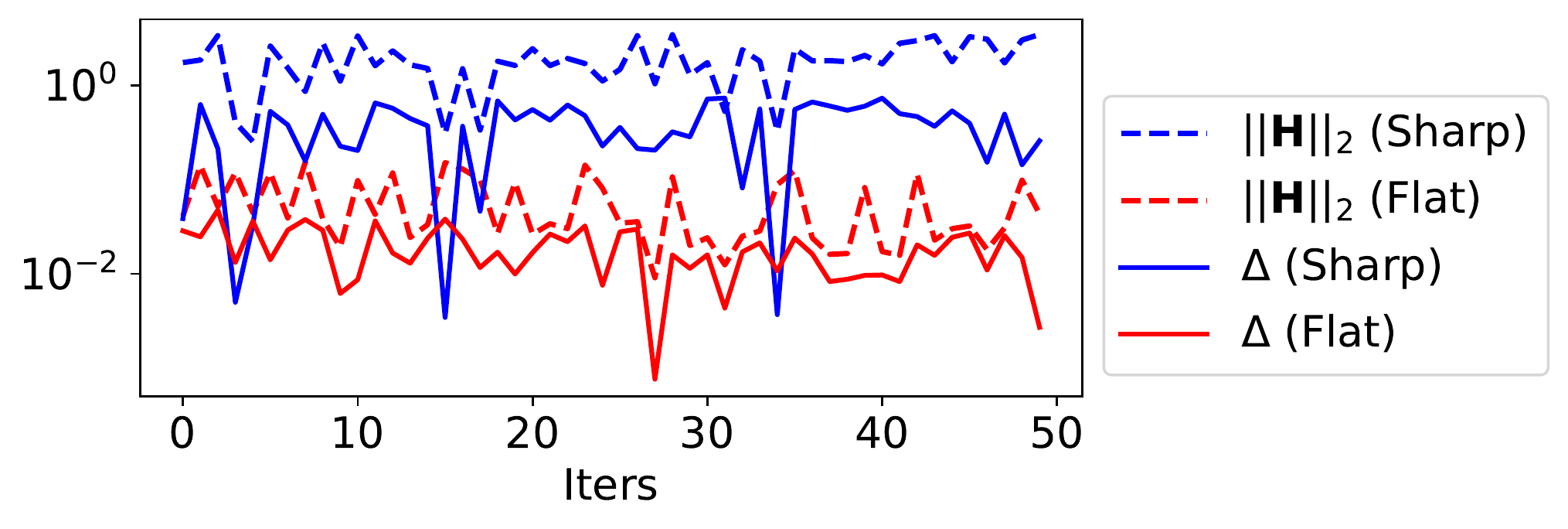}
    \caption{\textbf{(dashed line)} Sharpness $\|\mathbf{H}\|_2$ and \textbf{(solid line)} the difference of gradients after updating
without noise $\Delta$ on toy example of \Figref{fig:loss_landscape}. The dashed and solid lines in the same color show a positive correlation during the training. Small $\|\mathbf{H}\|_2$ in flat loss surface (red) bounds the gradient difference $\Delta$ compared to the sharp one (blue).}
    \label{fig:toy_flatsharp}
\end{figure}

To push further, we investigate the relationship between sharpness and noise addition. Motivated by \cite{shamsabadi2023losing}, the following theorem illustrates that flat minima can reduce the error caused by adding Gaussian noise to the clipped gradient in \Eqref{eq:DP-SGD}. 
\begin{theorem}\label{thm:noise}
\textbf{(Flat minimum reduces the bias of noise addition)} The difference of gradients between updated without noise $\bar \vg_t$ and with noise $\vg_t$ is affected by the sharpness as  
$$
\|\nabla \ell(\vw_t-\eta\bar\vg_t) - \nabla \ell(\vw_t-\eta\vg_t) \|\leq \eta \max_{\mathbf{H} \in \mathbb{H}}(\|\mathbf{H}\|_2) \cdot\|\boldsymbol{\mu}\|,
$$

where $\boldsymbol{\mu} \sim \mathcal{N}(\mathbf{0},C^2\sigma^2\mathbf{I})$ and $\mathbb{H}$ is a set of Hessian matrices $\mathbf{H}$ along the line of $\boldsymbol{\mu}$ from $\vw_t-\eta \bar\vg_t$ to $\vw_t-\eta \vg_t$.
\end{theorem}
We defer the proof to \Appref{app:noise}. Theorem \ref{thm:noise} suggests that the sharpness $\|\mathbf{H}\|_2$ can regulate the impact of Gaussian noise in the training, even with the same learning rate $\eta$, clipping value $C$, and noise level $\sigma$.  

We now empirically validate \Thmref{thm:noise}. Revisit \Figref{fig:loss_landscape} illustrating a toy example of a two-dimensional parameter space that comprises one sharp and one flat minimum, suggested in \cite{wang2021dplis}. Both the sharp and flat minimum have a loss value of 0. Please refer to \Appref{app:toy_setting} for the details.
In \Figref{fig:toy_flatsharp}, we measure the sharpness $\|\mathbf{H}\|_2$ and the gradient difference between updating with and without noise $\Delta_t := \|\nabla \ell(\vw_t-\eta\bar\vg_t) - \nabla \ell(\vw_t-\eta\vg_t) \|$ as training proceeds by gradient descents. To clearly show the effect of noise addition, we select initial points that are equidistant from each minimum and add the same level of noise in each step. 

The results show that a positive correlation exists between the sharpness $\|\mathbf{H}\|_2$ (dashed line) and the gradient difference $\Delta$ (solid line) during all training epochs, regardless of whether flat (red colored) or sharp (blue colored). More importantly, a flat minimum (red colored) has a lower value of the sharpness $\|\mathbf{H}\|_2$ (dashed line) and thus the gradient difference $\Delta$ (solid line), compared to the sharp one (blue colored).



\section{Discovering Flat Minima in Private Learning}
In the previous section, we show the importance of sharpness in private learning. To seek flat minima, we first demonstrate the effectiveness of SAM in DP training. However, at the same time, we also emphasize the drawbacks of SAM for private learning in terms of privacy budget and computational time. To address these limitations, we propose a new DP-friendly sharpness-aware training algorithm.
\subsection{Challenges of SAM in Private Learning}
\label{sec:dpsam}

\begin{figure}[!ht]
\centering     
    \subfigure[SGD and SAM]{\label{fig:loss_nondp}\includegraphics[width=40mm]{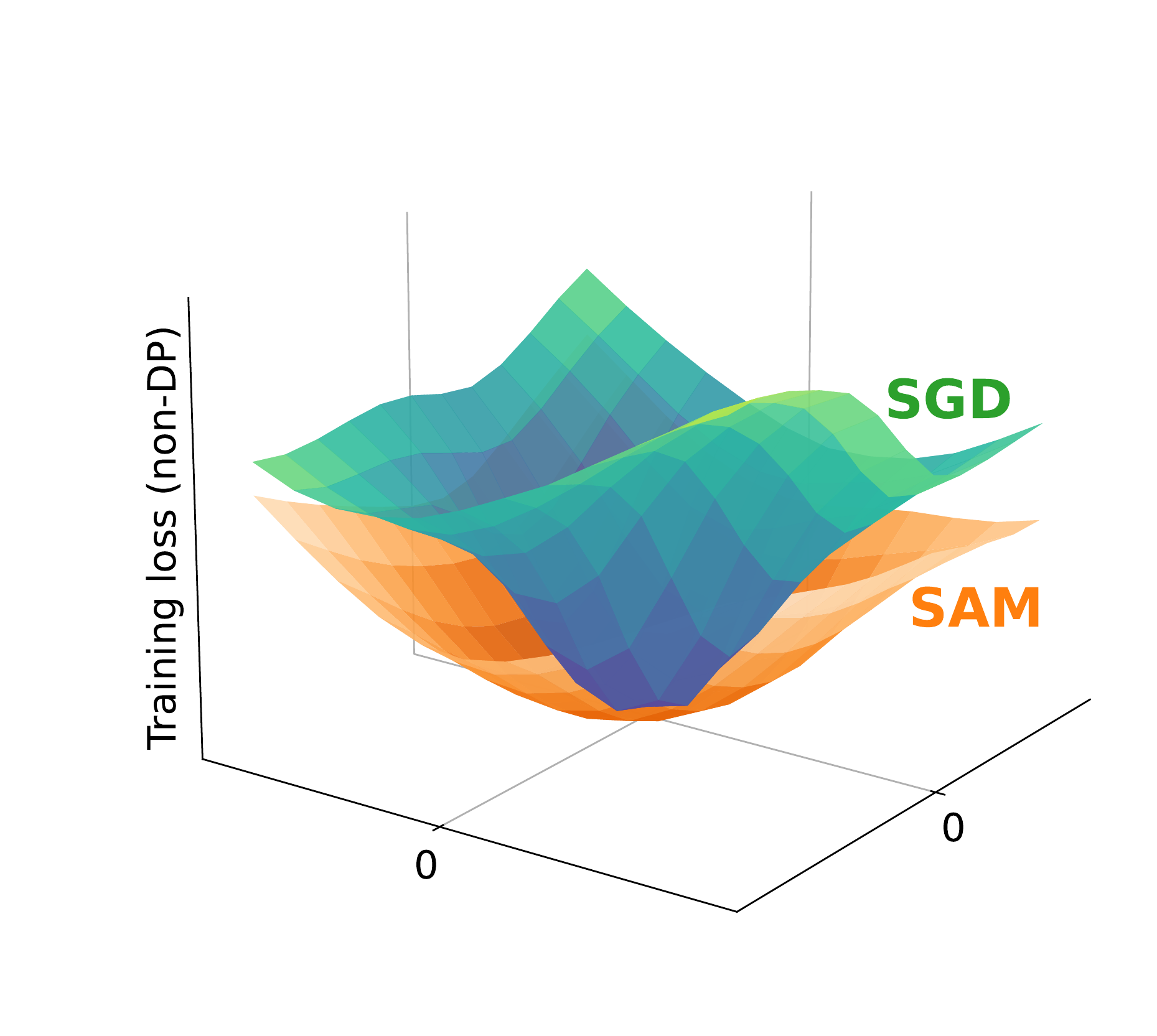}}
    \subfigure[DP-SGD and DP-SAM]{\label{fig:loss_dp}\includegraphics[width=40mm]{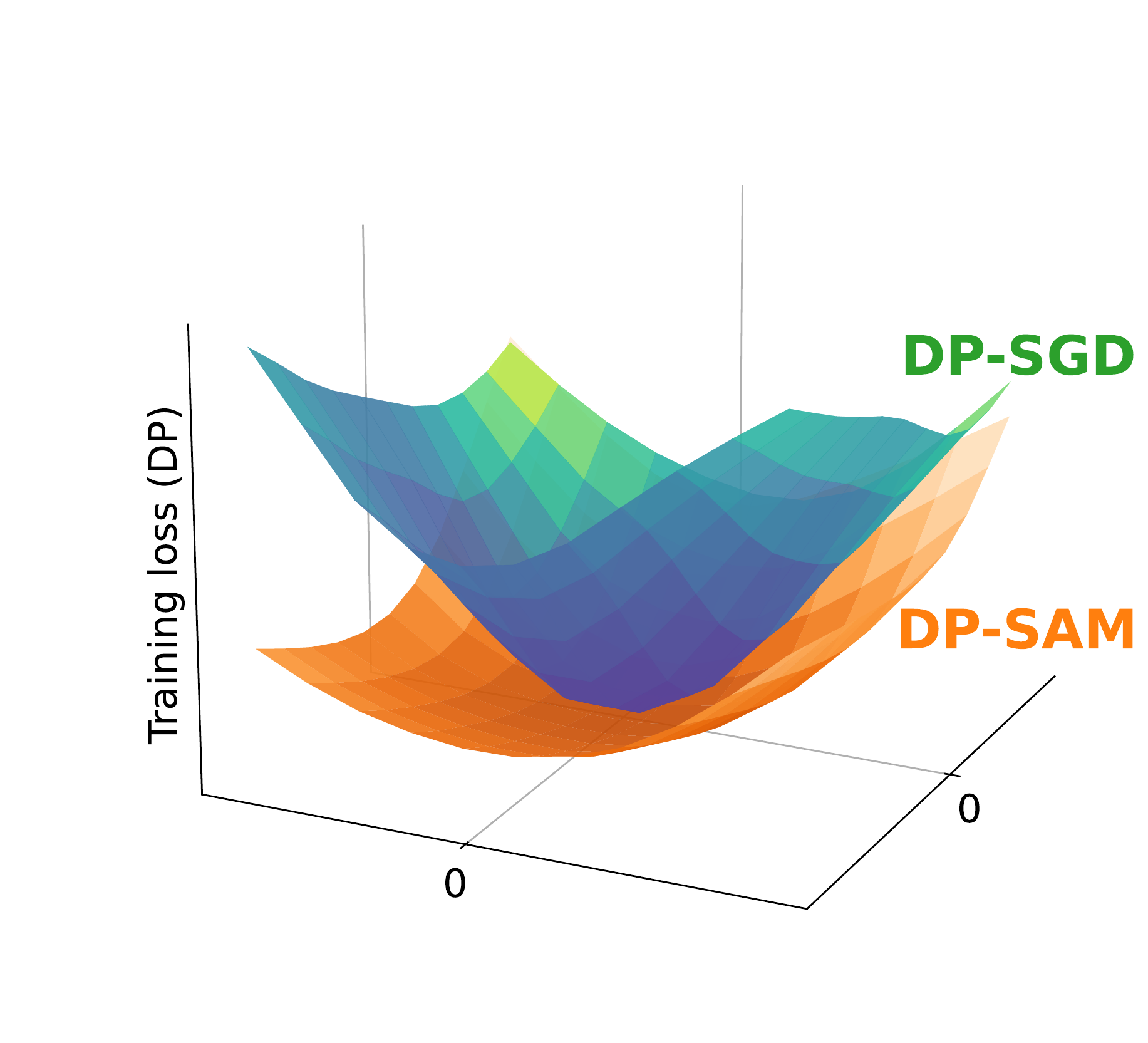}}
    \label{fig:losses_dp}
    \caption{Visualization of loss landscapes.}
\label{fig:losses}
\end{figure}

We introduce the concept of SAM for DP-SGD to achieve flat minima in DP training.
One of the main advantages of SAM is the ease of implementation, as it can be applied to various optimizers and architectures by modifying the gradient descent of SGD to a two-step optimization.  Thus, we can easily formulate SAM for DP training, referred to as \textit{DP-SAM}, as follows:
\begin{align}
    \vw_t^p &= \vw_t + \rho \frac{\vg_t}{\| \vg_t \|},  \label{eq:dpsam_ascent}\\
    \vg^{p}_t &= \frac{1}{|I_t|} \sum_{i\in I_t}  \text{clip}(\nabla\ell_i(\vw_t^p),C)+\mathcal{N}(\mathbf{0},C^2\sigma^2\mathbf{I}),
    \label{eq:dpsam_descent}  
\end{align}
where ${\vw}_{t+1} = {\vw_t} - \eta \vg^{p}_t$ and $\rho$ is the radius in the parameter space.  
In \Figref{fig:losses}, we use the techniques of \cite{li2018visualizing} to visualize the effectiveness of SAM on the loss landscape for non-DP (left) and DP training (right). Specifically, we perturb a converged minimum to two randomly sampled Gaussian directions and calculate all the losses in grids. The results demonstrate that DP-SAM is more effective than DP-SGD in uncovering flat minima in private learning, which is consistent with the results of standard training with SGD and SAM.

Nevertheless, we argue that DP-SAM may not be suitable for private learning, despite its ability to discover flat minima. For further details of Equations \ref{eq:dpsam_ascent} and \ref{eq:dpsam_descent}, we should consider that SAM is a two-step optimization that employs the data samples within the same mini-batch twice.  Therefore, we should inject noise into the gradients of the current weight $\vg_t$ and the perturbed weight $\vg^{p}_t$ to make both ascent and descent steps private. 
This might have two drawbacks, i.e., increased privacy cost and computational burden.

\paragraph{Increased privacy cost} Because every query to training data increases the privacy budget, we need to consider the privacy budget of the two-step optimization in Equations \peqref{eq:dpsam_ascent} and \peqref{eq:dpsam_descent}. We now demonstrate the privacy guarantee of DP-SAM. 
\begin{theorem}
\label{thm:dpsam}
(Privacy guarantee) 
DP-SAM requires $(2\varepsilon, 2\delta)$-differential privacy, whereas DP-SGD is $(\varepsilon,\delta)$-differential privacy.
\end{theorem}
\begin{proof} 
(Stated informally) For DP-SAM, let $\mathcal{M}_1(d)$ be the ascent step to calculate $\vw_t^p$ in \Eqref{eq:dpsam_ascent} and $\mathcal{M}_2(d,\mathcal{M}_1(d))$ be the descent step to calculate $\vg^p_t$ in \Eqref{eq:dpsam_descent}, $\forall d \in \mathcal{D}$. According to Proposition \ref{prop:dpsgd_noise}, $\mathcal{M}_1(d)$ and $\mathcal{M}_2(d)$ satisfy $(\varepsilon,\delta)$-DP by clipping individual gradients and injecting noise with $\sigma$. Then, by 
the general composition, DP-SAM requires the addition of two privacy budgets, resulting in ($2\varepsilon,2\delta)$-DP. 
The detailed mathematical proof can be found in \Appref{sec:proof_dpsam}.
\label{pf:dpsam}
\end{proof}

Note that the utilization of the moments accountant of Proposition \ref{prop:dpsgd_noise} \cite{abadi2016deep} or the advanced composition theorem \cite{dwork2014algorithmic} is not feasible for DP-SAM, as these theorems require the random selection of data, known as a \textit{k-fold composition experiment}.

To ensure the same level of privacy as DP-SGD, DP-SAM must satisfy either of the following conditions: $2\sqrt\frac{\log\delta}{\log2\delta}$ $(\approx 2.06 \text{ when } \delta=10^{-5})$ times the noise level $\sigma$ or $\frac{1}{4}\frac{\log 2\delta}{\log \delta}$ $(\approx 0.24 \text{ when } \delta=10^{-5})$ times the number of the training iterations compared to DP-SGD. In this paper, we choose to reduce the training time, which usually yields better performance than increasing the noise levels. 


\begin{figure*}[!ht]
\centering     
    \includegraphics[width=160mm]{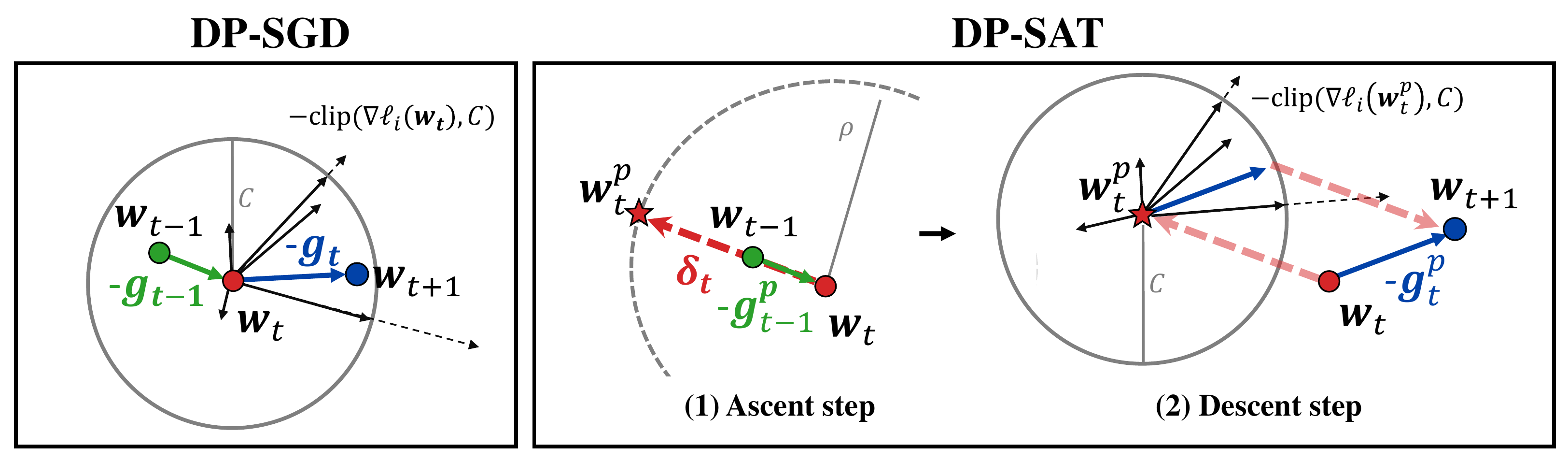}
    \caption{Illustration of DP-SGD and DP-SAT at step $t$. For ease of understanding, we set the learning rate $\eta=1$.}
    \label{fig:figure_algorithm_dp-sat}
\end{figure*}

\paragraph{Computational overhead}
The primary weakness of SAM is its computation overhead for implementing two-step optimization \cite{du2021efficient}. 
Furthermore, the computational burden of DP-SAM may impede the use of sharpness-aware training methods because DP training already has a large computational burden \cite{li2022large}.

Briefly, when given $m$ data samples with $d$-dimensional features, a model size of $w$, DP-SGD requires $O(mdw)$ for one forward and backward step. However, DP-SAM requires the doubled computations of $O(mdw)$ because it requires two times of gradient computations. Note that clipping operation for DP training is  generally $O(mw)$ and storing and recovering weights or individual gradients needs $O(w)$ computation.

\subsection{DP-SAT: Differentially Private Sharpness-Aware Training}
\label{sec:dpsat}

As aforementioned, DP-SAM yields a privacy-optimization trade-off, which results in a decrease in classification performance while attaining flat minima. 
The primary problem of DP-SAM is that it requires twice of privacy budget compared to DP-SGD.
To maintain the privacy budget consumption of DP-SGD in each iteration, we should abstain from accessing the training data multiple times in a mini-batch. 

To achieve this goal, we make use of post-processing \cite{dwork2014algorithmic}. Because post-processing guarantees that prior differentially private outputs do not impact the privacy budget, the use of the perturbed gradients is permitted in the earlier steps $\vg^p_{1},\ldots,\vg^p_{t-1}$ without any cost. 

\begin{algorithm}[t]\DontPrintSemicolon
\SetAlgoLined
\SetNoFillComment
    \caption{DP-SAT}
    \label{alg:main}
    \KwIn {Initial parameter $\vw_0$, learning rate $\eta$, radius $\rho$, clipping threshold $C$, variance $\sigma^2$ from Proposition \ref{prop:dpsgd_noise}, and small $\tau$ to prevent zero division. }
    \KwOut {Final parameter $\vw_{T}$.}
    \textbf{Initialize:} $\vg^p_0=\mathbf{0}$\\
    \For {$t=1,2,\ldots,T$}{
        Construct a random mini-batch $I_t$
        
        \uIf{DP-SGD}{
             $\textcolor[rgb]{0.101,0.255,0.635}{\vg_t} = \frac{1}{|I_t|}\sum_{i\in I_t}  \text{clip}(\nabla \ell_i(\vw_t),C)+\mathcal{N}(\mathbf{0},C^2\sigma^2 \mathbf{I})$
             
             ${\vw}_{t+1} = {\vw_t} - \eta \textcolor[rgb]{0.101,0.255,0.635}{\vg_t}$ 
        }
    
        \uIf{DP-SAT}{
            $\textcolor[rgb]{0.772,0.223,0.196}{\vdelta_t} = \rho \textcolor[rgb]{0.165,0.631,0.167}{\vg^p_{t-1}} / (\|\textcolor[rgb]{0.165,0.631,0.167}{\vg^p_{t-1}}\|+\tau)$  \tcp*{Post-processing}
        
            $\textcolor[rgb]{0.101,0.255,0.635}{\vg^p_t}$ = {\tcp*{Sharpness-aware training}} $\quad \frac{1}{|I_t|} \sum_{i\in I_t}  \text{clip}(\nabla\ell_i(\vw_t+\textcolor[rgb]{0.772,0.223,0.196}{\vdelta_t}),C)+\mathcal{N}(\mathbf{0},C^2\sigma^2 \mathbf{I})$ 
            
            ${\vw}_{t+1} = {\vw_t} - \eta \textcolor[rgb]{0.101,0.255,0.635}{\vg^p_t}$ 
        }
    }

\end{algorithm}
Based on this motivation, we present a new method that re-use the perturbed gradient of the previous step $t-1$ to alter the ascent direction of DP-SAM at the current step $t$. To be specific, its ascent step can be formulated as follows:
\begin{align}
    \vw^p_t = \vw_t + \rho \frac {\vg^p_{t-1}} {\|{\vg^p_{t-1}}\|}.
    \label{eq:dpsat}  
\end{align}
We call this approach \textit{Differentially Private Sharpness-Aware Training (DP-SAT)}.
The detailed training procedure of DP-SAT is explained in \Algref{alg:main} and \Figref{fig:figure_algorithm_dp-sat}.

Now, we prove that DP-SAT satisfies $(\varepsilon,\delta)$-DP, which consumes the same privacy budget as DP-SGD.

\begin{theorem}
\label{thm:dpsat}
(Privacy guarantee) 
DP-SAT in \Algref{alg:main} can guarantee $(\varepsilon,\delta)$-differential privacy.
\end{theorem}
\begin{proof}
    (Stated informally) It is sufficient to mention that $\vw^p_t$ at step $t$ is a result of the post-processing of $\vg^p_{t-1}$. Then, DP-SAT in \Algref{alg:main} guarantees $(\varepsilon,\delta)$-DP under $\sigma$ satisfying Proposition \ref{prop:dpsgd_noise}, because it only accesses the training data within the current batch $I_t$ once, which is the same as DP-SGD. The detailed mathematical proof can be found in \Appref{sec:proof_dpsat}.
\end{proof}

Moreover, the ascent step of DP-SAT in \Eqref{eq:dpsat} does not require additional forward and backward propagation, which requires $O(mdw)$, because it uses the previous weight vector to calculate $\vw^p_t$, requiring the marginal computation of $O(w)$. 
In \Tabref{tab:comparison}, we summarize the comparison of sharpness, privacy budgets, and computational costs of DP-SGD, DP-SAM, and DP-SAT.

\begin{table}[!t]
\centering
\caption{Comparison of sharpness of minima, privacy budget, and computation cost of DP-SGD \cite{abadi2016deep}, DP-SAM, and DP-SAT. The privacy budget is estimated by assuming DP-SGD is ($\varepsilon,\delta$)-DP. Computational cost is calculated w.r.t. DP-SGD (1$\times$).}
\label{tab:comparison}
\resizebox{\textwidth/2}{!}{%
    \begin{tabular}{c|ccc}
    \toprule
         Methods   & Minima    & Privacy budget                 & Computational cost   \\ \hline
    DP-SGD  & Sharp      & \underline{($\varepsilon,\delta$)-DP}       & \underline{1$\times$}          \\
    DP-SAM & \underline{Flat}  & ($2\varepsilon,2\delta$)-DP       & 2$\times$ (doubled)          \\
    DP-SAT  & \underline{Flat} &  \underline{($\varepsilon,\delta$)-DP}        & \underline{1$\times$}  
    \\\bottomrule
    \end{tabular}
}
\end{table}

\begin{table*}[ht]
\centering
\caption{Classification accuracy of DP-SGD, DP-SAM, and DP-SAT on the MNIST, FashionMNIST, CIFAR-10, and SVHN datasets. We also report the Error Reduction Rate (ERR) when trained with DP-SAT, in comparison to DP-SGD. We bold the highest average accuracy.} 
\label{tab:main}
\resizebox{0.9\textwidth}{!}{%
\begin{tabular}{cccrrrr}
\toprule
\multirow{2}{*}{Datasets}     & \multirow{2}{*}{Model}                                                                          & \multirow{2}{*}{\begin{tabular}[c]{@{}c@{}}Privacy budget $\varepsilon$\\ ($\delta=10^{-5}$)\end{tabular}} & \multicolumn{3}{c}{Optimizers}                                                       & \multicolumn{1}{c}{\multirow{2}{*}{\begin{tabular}[c]{@{}c@{}}ERR\\ (\%)\end{tabular}}} \\ \cline{4-6}
                              &                                                                                                 &                                                                                                            & \multicolumn{1}{c}{DP-SGD} & \multicolumn{1}{c}{DP-SAM} & \multicolumn{1}{c}{DP-SAT} & \multicolumn{1}{c}{}                                                                    \\ \hline
\multirow{6}{*}{MNIST}        & \multirow{3}{*}{\begin{tabular}[c]{@{}c@{}}GNResNet-10\\ (\#Params: 4.90M)\end{tabular}}        & $\varepsilon=1$                                                                                            & 95.15±0.17                 & 92.50±0.44                 & \textbf{96.00±0.21}        & 17.53\%                                                                                 \\
                              &                                                                                                 & $\varepsilon=2$                                                                                            & 96.68±0.27                 & 94.52±0.47                 & \textbf{97.35±0.14}        & 20.18\%                                                                                 \\
                              &                                                                                                 & $\varepsilon=3$                                                                                            & 97.30±0.14                 & 95.62±0.29                 & \textbf{97.83±0.10}        & 19.63\%                                                                                 \\ \cline{2-7} 
                              & \multirow{3}{*}{\begin{tabular}[c]{@{}c@{}}DPNAS-MNIST\\ (\#Params: 0.21M)\end{tabular}}        & $\varepsilon=1$                                                                                            & 97.77±0.13                 & 97.21±0.31                 & \textbf{97.96±0.08}        & 8.52\%                                                                                  \\
                              &                                                                                                 & $\varepsilon=2$                                                                                            & 98.60±0.06                 & 97.94±0.20                 & \textbf{98.71±0.09}        & 7.86\%                                                                                  \\
                              &                                                                                                 & $\varepsilon=3$                                                                                            & 98.70±0.12                 & 98.11±0.33                 & \textbf{98.93±0.02}        & 17.69\%                                                                                 \\ \hline
\multirow{6}{*}{FashionMNIST} & \multirow{3}{*}{\begin{tabular}[c]{@{}c@{}}GNResNet-10\\ (\#Params: 4.90M)\end{tabular}}        & $\varepsilon=1$                                                                                            & 80.57±0.25                 & 76.73±0.30                 & \textbf{81.33±0.45}        & 3.91\%                                                                                  \\
                              &                                                                                                 & $\varepsilon=2$                                                                                            & 82.71±0.35                 & 79.65±0.53                 & \textbf{84.53±0.41}        & 10.53\%                                                                                 \\
                              &                                                                                                 & $\varepsilon=3$                                                                                            & 84.55±0.17                 & 80.68±0.39                 & \textbf{85.91±0.22}        & 8.80\%                                                                                   \\ \cline{2-7} 
                              & \multirow{3}{*}{\begin{tabular}[c]{@{}c@{}}DPNAS-MNIST\\ (\#Params: 0.21M)\end{tabular}}        & $\varepsilon=1$                                                                                            & 84.62±0.19                 & 82.13±0.39                 & \textbf{85.92±0.35}        & 8.45\%                                                                                  \\
                              &                                                                                                 & $\varepsilon=2$                                                                                            & 86.99±0.57                 & 84.21±0.42                 & \textbf{87.75±0.24}        & 5.84\%                                                                                  \\
                              &                                                                                                 & $\varepsilon=3$                                                                                            & 87.97±0.17                 & 84.58±0.56                 & \textbf{88.60±0.04}        & 5.24\%                                                                                  \\ \hline
\multirow{6}{*}{CIFAR-10}     & \multirow{3}{*}{\begin{tabular}[c]{@{}c@{}}CNN-Tanh with SELU\\ (\#Params: 0.55M)\end{tabular}} & $\varepsilon=1$                                                                                            & 45.24±0.42                 & 44.30±1.16                 & \textbf{45.78±0.48}        & 0.99\%                                                                                  \\
                              &                                                                                                 & $\varepsilon=2$                                                                                            & 56.90±0.33                 & 51.32±0.34                 & \textbf{58.35±0.55}        & 3.36\%                                                                                  \\
                              &                                                                                                 & $\varepsilon=3$                                                                                            & 61.84±0.48                 & 51.81±0.62                 & \textbf{63.51±0.40}        & 4.38\%                                                                                  \\ \cline{2-7} 
                              & \multirow{3}{*}{\begin{tabular}[c]{@{}c@{}}DPNAS-CIFAR10\\ (\#Params: 0.53M)\end{tabular}}      & $\varepsilon=1$                                                                                            & 59.42±0.38                 & 54.00±0.84                 & \textbf{60.13±0.34}        & 1.75\%                                                                                  \\
                              &                                                                                                 & $\varepsilon=2$                                                                                            & 66.30±0.27                 & 60.38±0.46                 & \textbf{67.23±0.12}        & 2.76\%                                                                                  \\
                              &                                                                                                 & $\varepsilon=3$                                                                                            & 68.43±0.43                 & 61.51±0.39                 & \textbf{69.86±0.49}        & 4.53\%                                           \\ \hline
              \multirow{3}{*}{SVHN}     & \multirow{3}{*}{\begin{tabular}[c]{@{}c@{}}DPNAS-CIFAR10\\ (\#Params: 0.53M)\end{tabular}} & $\varepsilon=1$                                                                                            & 82.25±0.15                 & 80.64±0.34            & \textbf{83.09±0.54}        & 4.73\%                                                                                  \\
                              &                                                                                                 & $\varepsilon=2$                                                                                            & 86.85±0.33	               & 85.06±0.26               & \textbf{87.68±0.13}        & 6.31\%                                                                                  \\
                              &                                                                                                 & $\varepsilon=3$                                                                                            & 88.18±0.23               & 86.24±0.22                 & \textbf{88.74±0.18}        & 4.74\%                                             
\\ \bottomrule
\end{tabular}%
}
\end{table*}

\paragraph{Higher gradient similarities in DP training}
The utilization of previous gradients in DP training is facilitated by employing a small clipping value $C$ and a larger batch size, distinguishing it from standard training. This choice leads to increased gradient similarities between the current and previous steps. Nevertheless, the process of finding an appropriate ascent step still poses challenges \cite{andriushchenko2022towards}. A comprehensive explanation is provided in \Appref{app:sat_motivation}.

\paragraph{Momentum variants}
Recent studies have examined the potential benefits of using momentum variants to enhance generalization and determine the optimal ascent step for SAM \cite{du2022sharpnessaware,park2023efficient}. 
Similarly, DP-SAT enables the utilization of all the previous step's privatized gradients $\vg_1,\ldots,\vg_{t-1}$, similar to the aforementioned momentum-based approaches. Our empirical investigation confirms that using the previous gradient alone is sufficient to achieve meaningful performance enhancements, similar to the momentum approach. We believe that the introduction of noise in the DP training process may impede the effective utilization of momentum. We refer the readers to \Appref{app:momentum} for detailed experiments. 

\section{Experiments}
\label{sec:exp}

\subsection{Experimental Setup}
For the empirical results trained from scratch, we evaluate the performance of our method on three commonly used benchmarks for differentially private deep learning: MNIST, FashionMNIST, CIFAR-10, and SVHN.
For architecture, we select various architectures: GNResNet-10 (Group Norm ResNet-10) and DPNASNet-MNIST \cite{cheng2022dpnas} for MNIST and FashionMNIST CNN-Tanh \cite{papernot2021tempered} with SELU and DPNASNet-CIFAR \cite{cheng2022dpnas} for CIFAR-10, and also DPNASNet-CIFAR for SVHN. Particularly, DPNASNet architectures are state-of-the-art architectures with DP-SGD from scratch.

Due to the serious accuracy drop for private learning from scratch, recent studies explore the use of fine-tuning and transfer learning in natural language processing \cite{yu2021large,yu2022differentially,li2022large} and computer vision \cite{bu2022scalable}. For fine-tuning, we evaluate various pre-trained Vision Transformers (ViT), such as  ViT \cite{dosovitskiy2020image}, DeiT \cite{touvron2021training}, and CrossViT \cite{chen2021crossvit}, with a wide range of model parameters using mixed ghost clipping proposed in \cite{bu2022scalable} for CIFAR-10 and CIFAR-100. 

The training data for each dataset was partitioned into training and test sets with a ratio of 0.8:0.2, and the test accuracy was averaged over 5 different random seeds for each dataset. All experiments are conducted using the PyTorch-based libraries \cite{kim2020torchattacks,opacus} with Python on four NVIDIA GeForce RTX 3090 GPUs. Please refer to \Appref{app:exp_setting} for more details of experimental settings. 

\begin{figure*}[!ht]
\centering     
    \includegraphics[width=160mm]{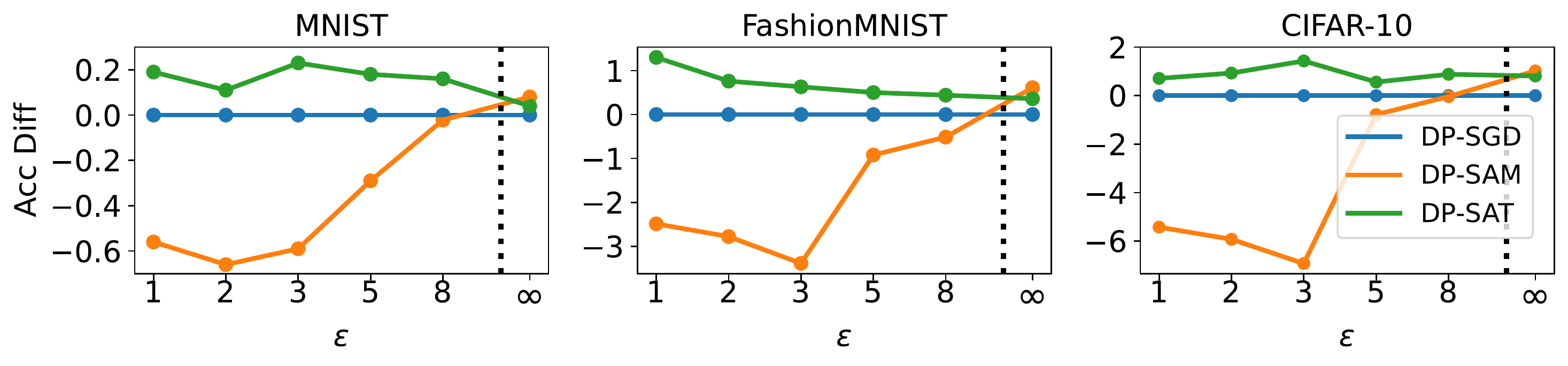}
    \caption{Difference of accuracy for DP-SAM and DP-SAT w.r.t DP-SGD. DPNAS architectures are used. $\varepsilon=\infty$ indicates non-DP settings.}
    \label{fig:acc_dpnondp}
\end{figure*}

\begin{table*}[ht!]
\centering
\caption{Fine-tuning accuracy of DP-SGD and DP-SAT on the CIFAR-10 and CIFAR-100 datasets with various pre-trained models. We also report the Error Reduction Rate (ERR) when trained with DP-SAT, in comparison to DP-SGD. The bold indicates results within the standard deviation of the highest mean score.} 
\label{tab:fine_tuning}
\resizebox{\textwidth}{!}{%
\begin{tabular}{c|cc|crc|crc}
\toprule
\multirow{3}{*}{\begin{tabular}[c]{@{}c@{}c@{}}Privacy\\ Budget $\varepsilon$\\ $(\delta=10^{-5})$ \\\end{tabular}} & \multicolumn{2}{c|}{Datasets} & \multicolumn{3}{c|}{CIFAR-10} & \multicolumn{3}{c}{CIFAR-100} \\ \cline{2-9} 
 & \multirow{2}{*}{Model} & \multirow{2}{*}{\#Params} & \multicolumn{2}{c|}{Optimizers} & \multicolumn{1}{c|}{\multirow{2}{*}{\begin{tabular}[c]{@{}c@{}}ERR\\ (\%)\end{tabular}}} & \multicolumn{2}{c|}{Optimizers} & \multicolumn{1}{c}{\multirow{2}{*}{\begin{tabular}[c]{@{}c@{}}ERR\\ (\%)\end{tabular}}} \\ \cline{4-5} \cline{7-8}
 &  &  & \multicolumn{1}{c}{DP-SGD} & \multicolumn{1}{c|}{DP-SAT} & \multicolumn{1}{c|}{} & \multicolumn{1}{c}{DP-SGD} & \multicolumn{1}{c|}{DP-SAT} & \multicolumn{1}{c}{} \\ \hline
\multirow{3}{*}{$\varepsilon=0.5$} & CrossViT\_18\_240 & 42.6M & 93.63±0.14 & \multicolumn{1}{r|}{\textbf{93.90±0.23}} & 4.24\% & 66.15±0.42 & \multicolumn{1}{r|}{\textbf{66.97±0.49}} & 2.42\% \\
 & ViT\_small\_patch16\_224 & 85.8M & 89.94±0.19 & \multicolumn{1}{r|}{\textbf{90.21±0.20}} & 2.68\% & 37.37±1.13 & \multicolumn{1}{r|}{\textbf{39.81±1.59}} & 3.9\% \\
 & DeiT\_base\_patch16\_224 & 85.8M & \textbf{92.21±0.24} & \multicolumn{1}{r|}{\textbf{92.15±0.15}} & -0.77\% & 49.25±1.03 & \multicolumn{1}{r|}{\textbf{50.03±0.59}} & 1.54\% \\ \hline
\multirow{5}{*}{$\varepsilon=2$} & CrossViT\_tiny\_240 & 6.7M & 88.34±0.25 & \multicolumn{1}{r|}{\textbf{88.76±0.28}} & 3.60\% & \textbf{59.58±0.31} & \multicolumn{1}{r|}{\textbf{59.75±0.40}} & 0.42\% \\
 & CrossViT\_small\_240 & 26.3M & 93.89±0.18 & \multicolumn{1}{r|}{\textbf{94.15±0.28}} & 4.26\% & 71.14±0.38 & \multicolumn{1}{r|}{\textbf{71.38±0.23}} & 0.83\% \\
 & CrossViT\_18\_240 & 42.6M & \textbf{95.32±0.12} & \multicolumn{1}{r|}{\textbf{95.31±0.13}} & -0.21\% & 74.29±0.17 & \multicolumn{1}{r|}{\textbf{74.52±0.19}} & 0.89\% \\
 & ViT\_small\_patch16\_224 & 85.8M & 92.22±0.46 & \multicolumn{1}{r|}{\textbf{92.50±0.27}} & 3.60\% & 65.89±0.72 & \multicolumn{1}{r|}{\textbf{67.27±0.67}} & 4.05\% \\
 & DeiT\_base\_patch16\_224 & 85.8M & \textbf{94.29±0.18} & \multicolumn{1}{r|}{\textbf{94.44±0.25}} & 2.63\% & \textbf{69.20±0.49} & \multicolumn{1}{r|}{\textbf{69.69±0.76}} & 1.59\%
\\ \bottomrule
\end{tabular}%
}
\end{table*}

\subsection{Classification Performance}
We conducted a performance comparison of DP-SGD, DP-SAM, and DP-SAT as presented in \Tabref{tab:main}. The proposed DP-SAT exhibits superior classification performance compared to DP-SGD in all scenarios, including both small and large models.
Specifically, DP-SAT enhances performance under DP-friendly architectures with fewer parameters, including state-of-the-art models such as DPNASNet architectures and CNN-Tanh, achieving an  (ERR) of 5.81\% on average in these scenarios. Moreover, the performance improvements are particularly pronounced in large models, such as GNResNet-10, with an ERR of 13.43\% on average.
As large models with complicated architectures are known to face challenges in generalizing well in DP settings due to their susceptibility to perturbations \cite{tramer2021differentially}, the identification of flat minima becomes notably advantageous.
Meanwhile, due to the aforementioned privacy consumption, DP-SAM shows lower classification performance than DP-SGD. 

To visualize the difference between optimization methods, we plot the accuracy difference with respect to DP-SGD in \Figref{fig:acc_dpnondp}. We tested on various privacy budgets $\varepsilon \in \{1,2,3,5,8\}$, including the non-DP ($\varepsilon=\infty$) setting. Consistent with prior results, DP-SAT shows the best performance among methods. Interestingly, as the privacy budget $\varepsilon$ increases, gradually approaching the non-DP settings, the gap between DP-SAT and DP-SAM diminishes. Thus,  it is clear that DP-SAT fully utilizes the positive effects of flatness in DP models by evaluating a broad range of privacy budgets $\varepsilon$.


For ablation studies, we conduct a range of experiments, including a sensitivity analysis on the parameter $\rho$, a comparison of various base optimizers, and investigations into other relevant factors. We further argue that the accuracy improvement achieved by DP-SAT is not solely reliant on the enlarged hyperparameter search space. The detailed results and analysis can be found in \Appref{app:ablation}.

\subsection{Fine-tuning Performance}
We now show that the idea of sharpness-aware training is effective in fine-tuning for private models. 
As the ViT models have well-generalizing latent space, they show higher fine-tuning accuracy than other CNN models.
We tested DP-SGD and DP-SAM for various pre-trained ViT models on $\varepsilon=0.5$ and $2$, as shown in \Tabref{tab:fine_tuning}.
The experimental results show that DP-SAT outperforms DP-SGD in the majority of fine-tuning cases. Note that the difference in fine-tuning experiments is marginal compared to from-scratch training because of the relatively small training epoch of only 5 epochs.

\subsection{Sharpness Analysis}
\begin{figure}[t!]
\centering     
    \includegraphics[width=80mm]{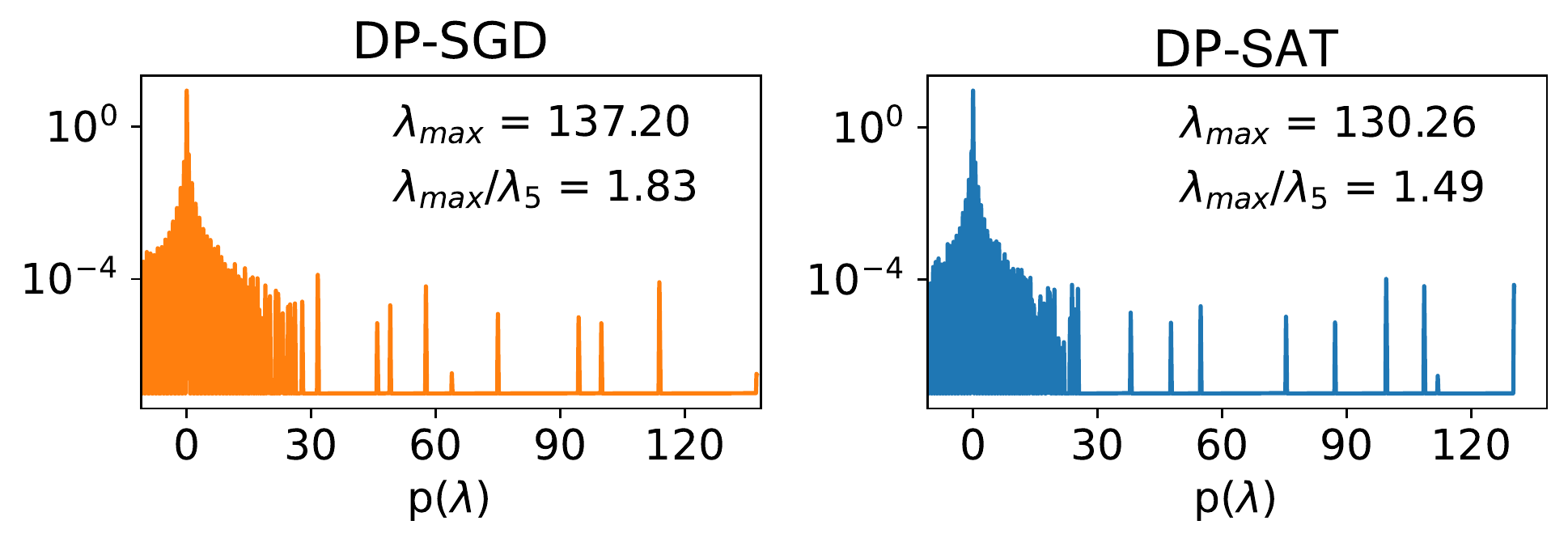}
    \caption{Eigenspectrum of Hessian on CIFAR-10.}
    \label{fig:hessian}
\end{figure}

We measure the Eigenspectrum of the Hessian matrix $\mathbf{H}$ of the trained models with DP-SGD and DP-SAT on CIFAR-10 in \Figref{fig:hessian}. In DP-SAT, the probability of eigenvalues $p(\lambda)$ is shifted towards the left, which indicates DP-SAT finds flatter minima compared to DP-SGD. In addition, both the sharpness $\lambda_{max}=130.26$ and the ratio of eigenvalues $\lambda_{max}/\lambda_5=1.49$, which are the popular measures to estimate flat minima, are smaller than those of DP-SGD.

\begin{figure}[t!]
\centering     
    \subfigure[Loss]{\label{fig:lineconnecting_loss}\includegraphics[width=40mm]{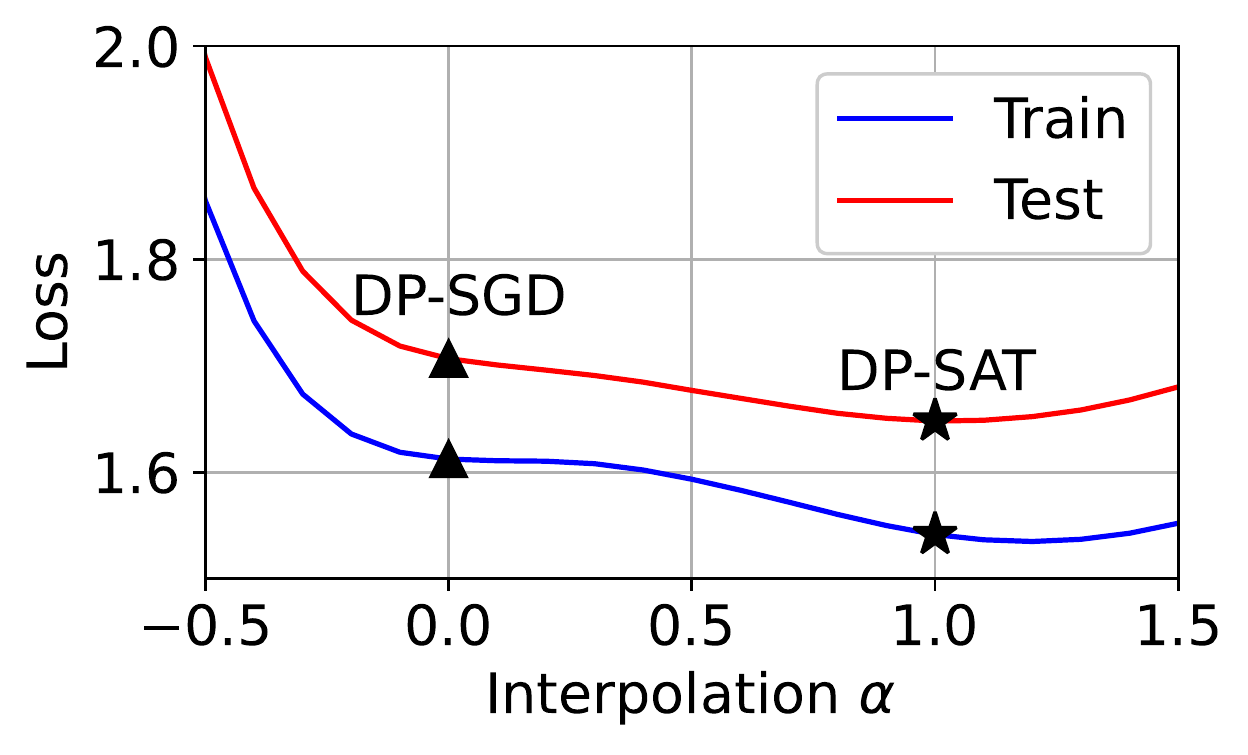}}
    \subfigure[Accuracy]{\label{fig:lineconnecting_acc}\includegraphics[width=40mm]{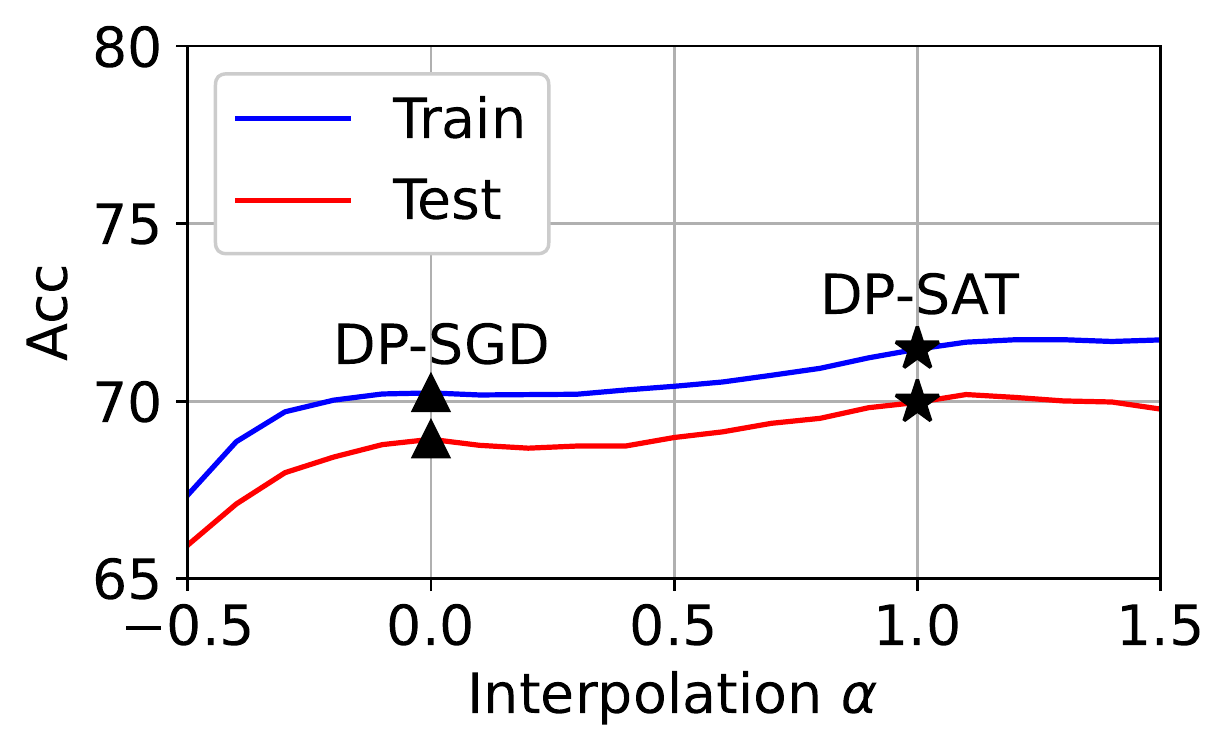}}
    \caption{Results of linear interpolations $\vw(\alpha) = (1 - \alpha)\vw + \alpha \vw'$ (for $\alpha \in [-0.5,1.5])$ between DP-SGD ($\blacktriangle$) and DP-SAT ($\star$).}
\label{fig:lineconnecting}
\end{figure}

Furthermore, we adopt the idea of \cite{chen2022when} to explain the success of DP-SAT, which can be explained by the similarities between training and test losses. Higher levels of similarity lead to a smaller generalization gap in performance under flatter local minima. This is a distinct factor contributing to the success of SAM that is separate from the impact of gradient norm.
We interpolate the loss surface of models trained using both DP-SGD and DP-SAT in \Figref{fig:lineconnecting}. Our results are consistent with well-generalizing settings and demonstrate that: (1) DP training exhibits higher levels of similarity between training and test losses; (2) DP-SAT produces a flatter loss landscape and lower errors in both training and test settings; and (3) the shape of the functions is strongly correlated with generalization performance.

\subsection{Computational Efficiency}
We empirically show that DP-SAT does not need an additional computational burden for sharpness-aware training in \Tabref{tab:comp_time}. The training speed of DP-SGD and DP-SAT is almost the same in all datasets. On the other hand, DP-SAM requires further training times to calculate the ascent direction at every step $t$. Note that the training speed ratio compared to DP-SGD is twice in non-DP training, but it shows lesser results such as 164.9\% or 181.9\% because DP training itself needs additional computation, such as individual gradient accumulation and memory access.

\begin{table}[!t]
\centering
\caption{Training speed (images/sec) on the MNIST, FashionMNIST, and CIFAR-10 (higher is faster). The numbers in parentheses ($\cdot$) indicate the training speed ratio w.r.t. DP-SGD (lower is faster).}
\label{tab:comp_time}
\resizebox{\textwidth/2}{!}{%
\begin{tabular}{c|lll}
\toprule
 & \multicolumn{1}{c}{MNIST} & \multicolumn{1}{c}{FashionMNIST} & \multicolumn{1}{c}{CIFAR-10} \\ \hline
DP-SGD     & 5,776            & 6,020           & 3,233           \\
DP-SAM     & 3,503 (164.9\%) & 3,712 (162.2\%) & 1,777 (181.9\%) \\
DP-SAT & 5,777 (100.0\%)   & 6,022 (100.0\%)  & 3,188 (101.4\%)
\\ \bottomrule 
\end{tabular}
}
\end{table}

\section{Limitations and Future Work}
\label{sec:limit}
As the relationship between flat minima and generalization performance is still being actively researched, we hope that further work will be built on our work to explore the advantages of flatness under DP training schemes. First, there is a distinct line of work that attempts to achieve flatness, referred to as weight averaging, which has been actively compared and combined with SAM in recent studies \cite{kaddour2022flat}. At present, it appears that weight averaging has difficulty in improving the performance of DP models \cite{panda2022dp}; however, we believe that suitable variants of weight averaging may be beneficial for DP training methods from our benchmark results. Second, research into gradient norm regularization under DP schemes could be a beneficial direction. Lastly, we will further investigate the recently proposed variants of SAM in DP training.

\section{Conclusion}
\label{sec:conclusion}
In this paper, we investigated the geometric properties of private learning, specifically sharpness. We showed that seeking flat minima can mitigate the negative effects of clipping and noise addition during training.
However, we also identified that the two-step optimization of SAM may have negative impacts on privacy budget and computational time. To address this issue, we proposed a new sharpness-aware training method that can improve performance without additional privacy or computational burden. We believe that this work will contribute to the understanding of sharpness and optimization in deep learning with differential privacy.

\section*{Acknowledgements}
This work was partly supported by the Institute of Information \& communications Technology Planning \& Evaluation (IITP) grant funded by the Korea government (MSIT) (No.2022-0-00984, Development of Artificial Intelligence Technology for Personalized Plug-and-Play Explanation and Verification of Explanation; No.2021-0-02068, Artificial Intelligence Innovation Hub) 
and the National Research Foundation of Korea (NRF) Grant funded by the Korean Government (MSIT) (No. 2019R1A2C2002358; No. 2022R1A5A600ff0840). 

\clearpage





\bibliography{ms}
\bibliographystyle{icml2023}

\newpage
\appendix
\onecolumn
\section{Composition Theorems}
\label{app:dp_properties}
We present the basic composition theorems of $(\varepsilon,\delta)$-DP algorithms.

\paragraph{General composition theorem \cite{dwork2014algorithmic}}

\begin{definition}\label{def:general_composition}
    (General composition theorem for $(\varepsilon,\delta)$-DP algorithms) 
    Let $\mathcal{M}_1: d\mapsto\mathcal{M}_1(d)\in\mathcal{R}_1$ be an $(\varepsilon,\delta)$-DP function, and for $k\geq2$ and $s_j \in \mathcal R_j ,\forall j \in \{1,\ldots, k-1 \}, \mathcal{M}_k: (d,s_1,\ldots,s_{k-1})\mapsto \mathcal{M}_k(d)\in \mathcal{R}_k$ be $(\varepsilon,\delta)$-DP, given the previous outputs $s_1,\ldots, s_{k-1}\in \otimes_{j=1}^{k-1} \mathcal{R}_j$. Then, for all neighboring $d, d'$ and all $\mathcal{S}\subset \otimes_{j=1}^k \mathcal{R}_j$,
\begin{equation}\label{eq:general_composition}
        Pr[(\mathcal{M}_1,\ldots, \mathcal{M}_k)(d)\in \mathcal{S}] \leq e^{k\varepsilon} Pr[(\mathcal{M}_1,\ldots, \mathcal{M}_k)(d') \in \mathcal{S}]+k\delta.
        \end{equation}
    \end{definition}

Note that \Eqref{eq:general_composition} does not require any assumption of $d$, which can be used in \textit{DP-SAM}.

\paragraph{Advanced composition theorem \cite{dwork2014algorithmic}}
From now on, we need strong assumptions on $d$, i.e., \textit{$k$-fold composition experiment}, which is the repeated use of differentially private algorithms on different (random sampled) data that may nevertheless contain information of one individual. By assumption on $d$ on each step, the concrete sequences of mechanisms can guarantee a tighter privacy budget than \Eqref{eq:general_composition}.

\begin{definition}\label{def:advanced_composition}
    (Advanced composition theorem for $(\varepsilon,\delta)$-DP algorithms) 
    For all $\varepsilon,\delta,\delta'\geq 0$, the sequence of $k$-fold ($\varepsilon,\delta$)-DP mechanisms satisfies ($\varepsilon',k\delta+\delta'$)-DP, where $\varepsilon'=\sqrt{2k\ln(1/\delta')}\varepsilon+k\varepsilon(e^\varepsilon-1).$
\end{definition}

\paragraph{Moments accountant \cite{abadi2016deep}}
\textbf{(restated) }\Eqref{eq:DP-SGD} illustrates the weight update at step $t$ as follows:
\begin{align*}
    \vg_t &=\frac{1}{|I_t|} \sum_{i\in I_t}  \text{clip}(\nabla \ell_i(\vw_t),C)+\mathcal{N}(\mathbf{0},C^2\sigma^2),\\
    {\vw}_{t+1} &= {\vw_t} - \eta\vg_t,
\end{align*}
where \text{clip}$(\vu,C)$ projects $\vu$ to the $L_2$-ball with radius $C$.
\citet{abadi2016deep} proved that there exist constant $c_1$ and $c_2$ so that given total steps $T$ and sampling probability $q$, for any $\varepsilon<c_1 q^2T$, \Eqref{eq:DP-SGD} guarantee ($\varepsilon,\delta$)-DP, for any $\delta>0$ if we choose 
\begin{equation}
    \sigma \geq c_2 \frac{q\sqrt{T\log(1/\delta})}{\varepsilon}.
    \label{eq:DP-SGD_sigma}
\end{equation}

The composition of moments can reduce the accumulated privacy budget to $(O(q\varepsilon \sqrt T), \delta)$-DP. Detailed proof can be found in Appendix B of \cite{abadi2016deep}.

\paragraph{Post-processing}
Post-processing guarantees to use the previous differentially private outputs. 
\begin{definition}\label{def:post-processing}
(Post-processing \cite{dwork2014algorithmic})
If a mechanism $\mathcal{M}:\mathcal{X}\rightarrow \mathcal{R}_1$ is $(\varepsilon, \delta)$-DP, for any randomized mapping $h: \mathcal{R}_1 \rightarrow \mathcal{R}_2$, $h \circ \mathcal{M} : \mathcal{X}\rightarrow \mathcal{R}_2$ is at least $(\varepsilon, \delta)$-DP. 
\end{definition}

\section{Proofs}

\subsection{Proof of Theorem \ref{thm:clipping}}
\label{app:clipping}
\begin{proof}
With the first-order Taylor expansion of the gradients $\nabla\ell_i(\vw)$ for $\vw$ near a local minimum,\footnote{The second-order Taylor expansion of the loss function (and correspondingly, the first-order Taylor expansion of the gradient) is commonly employed to analyze properties in the vicinity of critical points in deep learning optimization \cite{zhao2022penalizing,xie2022adaptive}.}
\begin{align*}
\|\nabla\ell_i(\vw)\| &\approx \|\nabla\ell_i(\vw^*) + \mathbf{H}_{\vw^*}^T(\vw-\vw^*)\|\\
    \leq & \|\nabla\ell_i(\vw^*)\| + \|\mathbf{H}_{\vw^*}^T(\vw-\vw^*)\|\\
    = & \|\mathbf{H}_{\vw^*}^T(\vw-\vw^*)\| \\
    (&\because \text{at local minimum $\vw^*$, } \|\nabla\ell_i(\vw^*)\| = 0) \\
    \leq & \|\mathbf{H}_{\vw^*}\|_2\cdot\|\vw-\vw^*\|.
\end{align*}
Let us define the clip operation as follows:
$$\text{clip}(\nabla \ell_i(\vw),C)= \nabla \ell_i(\vw)\cdot\frac{1}{\max(1,\frac{\| \nabla \ell_i(\vw)\|}{C})}.$$
Then,
 \begin{align*}
    \|\nabla\ell_i(\vw)-&\text{clip}(\nabla \ell_i(\vw),C)\|  \\
    = &\begin{dcases*}
         \|\nabla\ell_i(\vw)\| - C  & if $  \|\nabla\ell_i(\vw)\| > C $,\\
         0 & otherwise. 
        \end{dcases*}\\
    = & \mathbbm{1}(\|\nabla\ell_i(\vw)\|>C)\cdot(\|\nabla\ell_i(\vw)\|-C) \\
    \leq & \mathbbm{1}(\|\mathbf{H}_{\vw^*}\|_2\cdot\|\vw-\vw^*\|>C)\cdot(\|\mathbf{H}_{\vw^*}\|_2\cdot\|\vw-\vw^*\|-C)\\
    = & \mathbbm{1}(\|\mathbf{H}_{\vw^*}\|_2\Delta_{\vw}>C)\cdot(\|\mathbf{H}_{\vw^*}\|_2\Delta_{\vw}-C).
  \end{align*}
where $\Delta_{\vw} = \|\vw-\vw^*\|$ and $\mathbbm{1}$ denotes an indicator function.   
\end{proof}

\subsection{Proof of \Thmref{thm:noise}}
We modify the proof of \cite{wang2021dplis,shamsabadi2023losing}, which indicates the effect of smoothness in terms of $\beta$-smoothness, to illustrate how the sharpness affects the Gaussian noise addition during training. 
\label{app:noise}

\begin{proof} 
Let $\mathbf{H}_\vw:=\nabla^2 \ell(\vw)$.
\begin{align*}
\|\nabla \ell(\vw_t-\eta\bar\vg_t) - \nabla \ell(\vw_t-\eta\vg_t)\| 
 &=  \eta\|(\int_0^1 \mathbf{H}_{(\vw_t-\eta\vg_t)+z\boldsymbol\mu}\boldsymbol\mu dz)\|\\
&(\because (\vw_t-\eta\bar\vg_t)-(\vw_t-\eta\vg_t)=\eta\boldsymbol\mu)\\
&\leq \eta \int_0^1 \| \mathbf{H}_{(\vw_t-\eta\vg_t)+z\boldsymbol\mu}\boldsymbol\mu \| dz\\
&\leq \eta \max_{\mathbf{H} \in \mathbb{H}}(\|\mathbf{H}\|_2) \cdot\|\boldsymbol\mu \|
\end{align*}
where $\boldsymbol{\mu} \sim \mathcal{N}(\mathbf{0},C^2\sigma^2\mathbf{I})$ and $\mathbb{H}$ is a set of Hessian matrices $\mathbf{H}$ along the line of $\boldsymbol{\mu}$ from $\vw_t-\eta \bar\vg_t$ to $\vw_t-\eta \vg_t$.
\end{proof}

\subsection{Proof of \Thmref{thm:dpsam}}
\label{sec:proof_dpsam}
Given $(\varepsilon,\delta)$ privacy budget of DP-SGD, let each privacy budget for the $t$-th step is $(\varepsilon_t,\delta_t)$, which represents the additional privacy budget for calculating $\boldsymbol{g}_t$. Then, it is enough to show that the $t$-th update of DP-SAM is $(2\varepsilon_t,2\delta_t)$-DP. 

The $t$-th update of DP-SAM can be decomposed by two mechanisms:
$$
\text{(ascent step) } w_t^p:=\mathcal{M_1}({w_t}, d)= {w_{t}} + \rho\frac{\boldsymbol{g}_t}{\|\boldsymbol{g}_t\|} \in \mathcal{R}_{1,t+1}$$
$$
\text{(descent step) } w_{t+1}:=\mathcal{M_2}({w_t^p},w_t, d)= {w_{t}} - \eta \boldsymbol{g}_t^p\in \mathcal{R}_{2,t+1}
$$
where $d \in \mathcal{D}$  and $\boldsymbol{g}_t^p$ defined by \Eqref{eq:dpsam_descent}. Note that the input of each mechanism is differentially private except for input data $d$. 
Calculating each $\boldsymbol{g}_t$ and $\boldsymbol{g}_t^p$ consumes the privacy budget of $(\varepsilon_t,\delta_t)$ same as DP-SGD.  Therefore, the additional cost of each mechanism is $(\varepsilon_t,\delta_t)$. 
Then, by the general composition theorem,  
$$\mathcal{M}_2:(w_t^p,w_t, d)\mapsto \mathcal{M}_2(w_t^p,w_t,d)\in \mathcal{R}_{2,t+1}$$
be $(2\varepsilon_t,2\delta_t)$ -DP for any $d\in \mathcal{D}$, $w_t \in\mathcal{R}_{2, t}$, and $w_t^p \in  \mathcal{R}_{1, t+1}$.
Therefore, the privacy budget of DP-SAM is $(2\varepsilon,2\delta)$.

\subsection{Proof of \Thmref{thm:dpsat}}
\label{sec:proof_dpsat}
Let us consider the same setting as the proof of \Thmref{thm:dpsam} (presented in \Appref{sec:proof_dpsam}). Then, it is enough to show that the $t$-th update of DP-SAT is $(\varepsilon_t,\delta_t)$-DP.  

The $t$-th update of DP-SAT can be decomposed by two mechanisms:
$$\text{(ascent step) } w_t^p:=\mathcal{M_1}([{w_t}], [w_{t-1}^p])= {w_{t}} + \rho\frac{\boldsymbol{g}^p_{t-1}}{\|\boldsymbol{g}^p_{t-1}\|} \in \mathcal{R}_{1,t+1}$$
$$\text{(descent step) } w_{t+1}:=\mathcal{M_2}([{w_t^p}],[w_t], d)= {w_{t}} - \eta \boldsymbol{g}_t^p\in \mathcal{R}_{2,t+1}$$
where $d \in \mathcal{D}$ , $[w_t] = \{w_1, \cdots, w_t\} \in \otimes_{j=1}^{t} \mathcal{R}_{2, j}$ and $[ w_t^p ] = \{w_1^p, \cdots, w_t^p\} \in \otimes_{j=1}^{t} \mathcal{R}_{1, j+1}$ and $\boldsymbol{g}_t^p$ defined by \Eqref{eq:dpsam_descent}. We used $[w_t]$ or $[w_t^p]$ instead of $w_t$ or $w_t^p$ to clearly indicate the accumulation of noise. For example, $\mathcal{M}_2$ requires the same input as  Theorem 4.1; $({w_t^p},w_t, d)$. 

Then, since $\mathcal{M_2}([w_{t-1}^p], [{w_{t-1}}], \cdot): \mathcal{D} \rightarrow\mathcal{R}_{2,t}$ is $(\varepsilon_{t-1},\delta_{t-1})$-DP mechanism and $\mathcal{M_1}(\cdot, [{w_{t-1}}],[w_{t-1}^p]):\mathcal{R}_{2,t}\rightarrow\mathcal{R}_{1,t+1}$ is a randomized mapping, the ascent step requires no additional privacy budget by post-processing.
Moreover, the descent step requires the same privacy budget as DP-SGD, the total additional privacy budget of $t$-th update is $(\varepsilon_t,\delta_t)$-DP, the same as DP-SGD.
Therefore, the privacy budget of DP-SAT is $(\varepsilon,\delta)$.



\section{Experimental settings}
\label{app:exp_setting}
\subsection{Classification}
We use SGD as a base optimizer with a momentum of 0.9 and a learning rate of 2.0, without any learning rate decay, as mentioned in \cite{cheng2022dpnas}. We conducted a hyperparameter search on $\rho=\{0.005,0.01,0.02,0.03,0.05,0.1\}$, and the privacy broken probability $\delta = 10^{-5}$ in DP training.

\begin{table*}[ht!]
\centering
\caption{Hyperparameters for training on MNIST, FashionMNIST, CIFAR-10, and SVHN.}
\label{tab:hyp2}
\resizebox{\textwidth}{!}{%
\begin{tabular}{lc|rr|rr|rr|r}
\toprule
\multicolumn{2}{l|}{Dataset} &
  \multicolumn{2}{c|}{MNIST} &
  \multicolumn{2}{c|}{FashionMNIST} &
  \multicolumn{2}{c|}{CIFAR-10} &\multicolumn{1}{c}{SVHN}\\ \hline
  \multicolumn{2}{l|}{Architecture}            & GNResNet-10  & DPNAS-MNIST  & GNResNet-10  & DPNAS-MNIST & CNN-Tanh with SELU  & DPNAS-CIFAR10 & DPNAS-CIFAR10   \\ \hline
\multicolumn{2}{l|}{Optimizer}            & SGD  & SGD  & SGD  & SGD & SGD  & SGD & SGD \\ 
\multicolumn{2}{l|}{Epoch}                & 40   & 40    & 40   & 40   & 30   & 30 & 30    \\
\multicolumn{2}{l|}{Batch size}           & 2048 & 2048  & 2048 & 2048 & 2048 & 2048 & 2048 \\
\multicolumn{2}{l|}{Learning rate $\eta$} & 2    & 2 & 2    & 2    & 2    & 2 & 2\\
\multicolumn{2}{l|}{Momentum $\beta$} &
 0.9 &
 0.9 &
  0.9 &
 0.9 &
 0.9 &
 0.9 &
 0.9  \\
\multicolumn{2}{l|}{Max grad norm $C$}    & 0.1  & 0.1   & 0.1  & 0.1  & 0.1  & 0.1  & 0.1  \\ \hline
\multicolumn{1}{l|}{\multirow{3}{*}{\begin{tabular}[c]{@{}l@{}}Radius\\ $\rho$\end{tabular}}} &
  $\varepsilon=1$ & 0.03
   & 0.03
   & 0.03
   & 0.05
   & 0.02
  & 0.01 
  & 0.05  \\
\multicolumn{1}{l|}{}  & $\varepsilon=2$  &   0.03   &   0.03    &  0.03     & 0.03  &  0.1 & 0.01  & 0.05  \\
\multicolumn{1}{l|}{}  & $\varepsilon=3$  &  0.03    &   0.03    &   0.03    & 0.02  &  0.05 & 0.01 & 0.03
\\ \bottomrule
\end{tabular}%
}
\end{table*}
Note that the best radius $\rho$ can be varied according to the randomized noise addition of each random seed due to the instability of DP training.
For a detailed explanation of DPNAS architectures in \cite{cheng2022dpnas}, please refer to their official GitHub code from \url{https://github.com/TheSunWillRise/DPNAS}.

\subsection{Fine-tuning}
We use Adam as a base optimizer with a learning rate of 0.002. We trained the model for 5 epochs with a batch size of 1000 and a mini-batch size of 100. 
Here, we use a hyperparameter search on a wide range of radius than the classification $\rho=\{0.001,0.005,0.01,0.05,0.1,0.5,1\}$, as we use a smaller learning rate. 
We use the mixed ghost clipping \cite{bu2022scalable} and their official GitHub code from \url{https://github.com/woodyx218/private_vision}.

\begin{table*}[ht!]
\centering
\caption{Radius $\rho$ for fine-tuning on $\varepsilon=0.5$.}
\label{tab:hyp3}
\resizebox{0.6\textwidth}{!}{%
\begin{tabular}{crrrrr}
\toprule
  & \multicolumn{1}{c}{CrossViT\_18\_240} & 
 \multicolumn{1}{c}{ViT\_small\_patch16\_224} &
 \multicolumn{1}{c}{DeiT\_base\_patch16\_224} \\\hline
CIFAR-10  & 0.05& 0.1&0.1  \\
CIFAR-100 & 0.05&0.2 & 0.2 
\\ \bottomrule
\end{tabular}%
}
\end{table*}

\begin{table*}[ht!]
\centering
\caption{Radius $\rho$ for fine-tuning on $\varepsilon=2$.}
\label{tab:hyp4}
\resizebox{0.9\textwidth}{!}{%
\begin{tabular}{crrrrr}
\toprule
 & \multicolumn{1}{c}{CrossViT\_tiny\_240} & \multicolumn{1}{c}{CrossViT\_small\_240} &  \multicolumn{1}{c}{CrossViT\_18\_240} & 
 \multicolumn{1}{c}{ViT\_small\_patch16\_224} &
 \multicolumn{1}{c}{DeiT\_base\_patch16\_224} \\ \hline
CIFAR-10 & 0.05 & 1 &  0.5 & 0.001 & 0.05 \\
CIFAR-100 & 0.001 & 0.001 &  0.01 & 0.001 & 0.05 
\\ \bottomrule
\end{tabular}%
}
\end{table*}

\subsection{Toy Example of \Figref{fig:loss_landscape}}
\label{app:toy_setting}
\Figref{fig:toy_flatsharp} illustrates the simple mixture of flat and sharp minima to investigate the effect of sharpness. Following \cite{wang2021dplis}, we generate the example as follows:

$$
\min_\vw  \left[ \mathcal{F}_{\vc_1}(\vw)\cdot \frac{\phi(\vw,\vc_1)}{\phi(\vw,\vc_1)+\phi(\vw,\vc_2)} +  \mathcal{F}_{\vc_2}(\vw)\cdot \frac{\phi(\vw,\vc_2)}{\phi(\vw,\vc_1)+\phi(\vw,\vc_2)} \right]
$$

where $\vc_1,\vc_2\in\mathbb{R}^2$ are two fixed centers for flat and sharp minima, respectively. We used $\vc_1=[2.5,2.5]$ and $\vc_2=[7.5,7.5].$
 $\phi(\vw,\vc)=e^{-{\|\vw-\vc\|}}$, 
$\mathcal{F}_{\vc_1}(\vw)= \texttt{Sigmoid} \left( \frac{\|\vw-\vc_1\|}{5}-\frac{5}{\|\vw-\vc_1\|}\right)$, and $\mathcal{F}_{\vc_2}(\vw)= \texttt{Sigmoid} \left( \frac{5\|\vw-\vc_1\|}{5}-\frac{5}{5\|\vw-\vc_1\|}\right)$.




\section{Illustration of DP-SGD, DP-SAM, and DP-SAT}
\begin{figure*}[!ht]
\centering     
    \includegraphics[width=150mm]{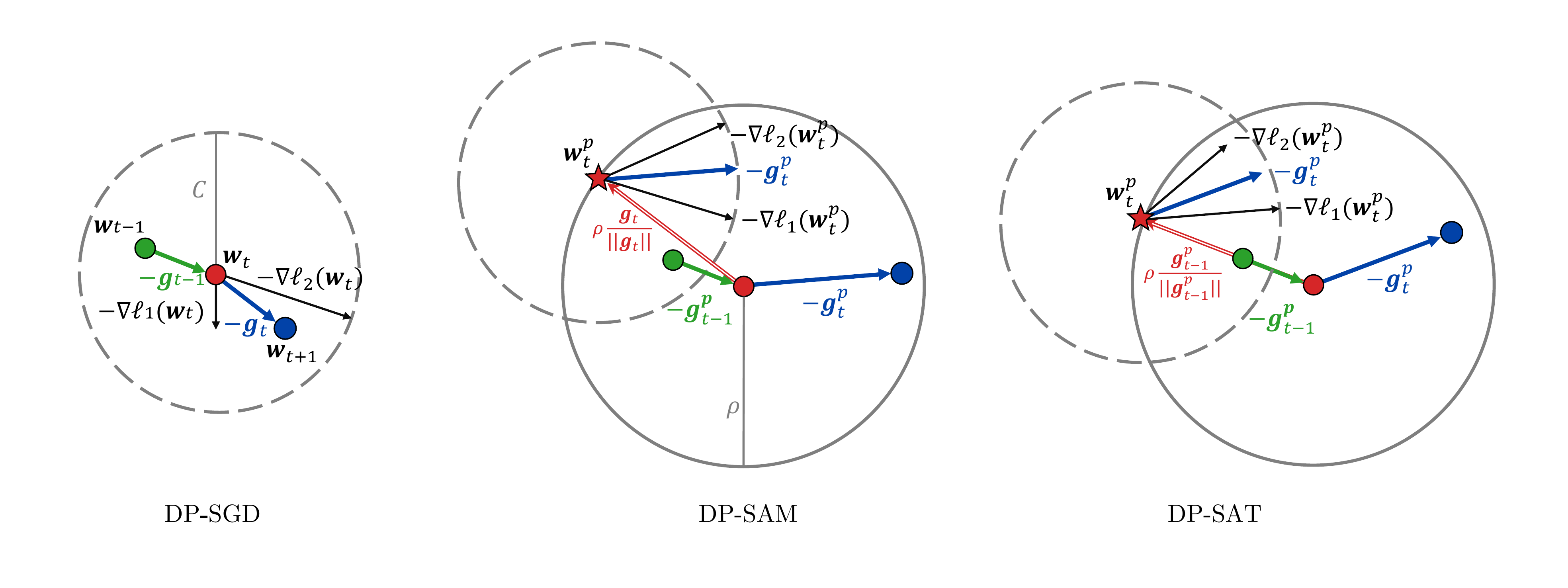}
    \caption{Illustration of DP-SGD, DP-SAM, and DP-SAT.}
    \label{fig:methods}
\end{figure*}

We illustrate the training methods of DP-SGD, DP-SAM, and DP-SAT in \Figref{fig:methods}.

We further compare the loss landscapes of DP-SGD, DP-SAM, and DP-SAT in the same way of \Figref{fig:losses}.
\begin{figure*}[!h]
\centering     
    \subfigure[DP-SGD]{\label{fig:sharpness_dp_dpsgd}\includegraphics[width=40mm]{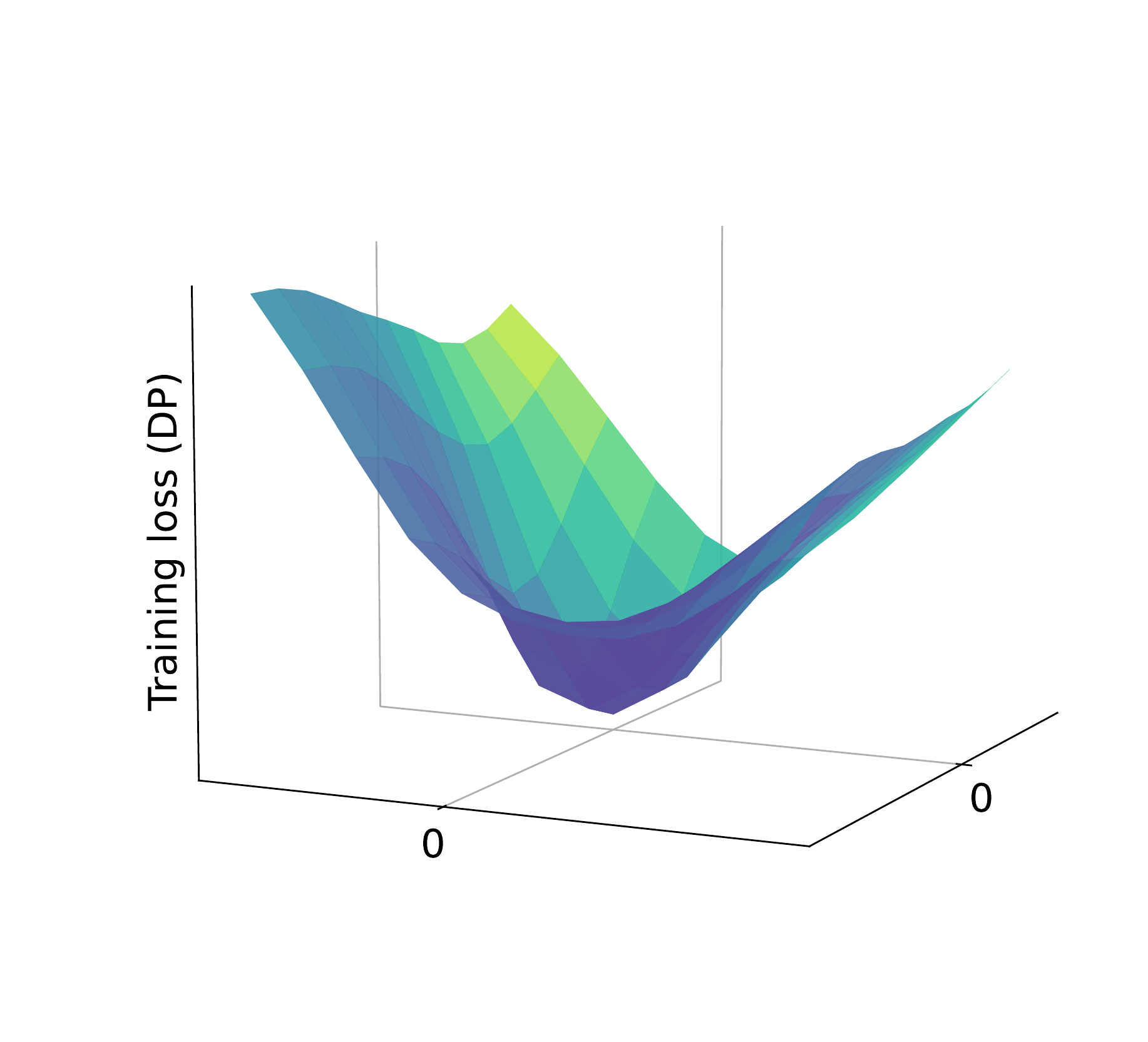}}
    \subfigure[DP-SAM]{\label{fig:sharpness_dp_dpsam}\includegraphics[width=40mm]{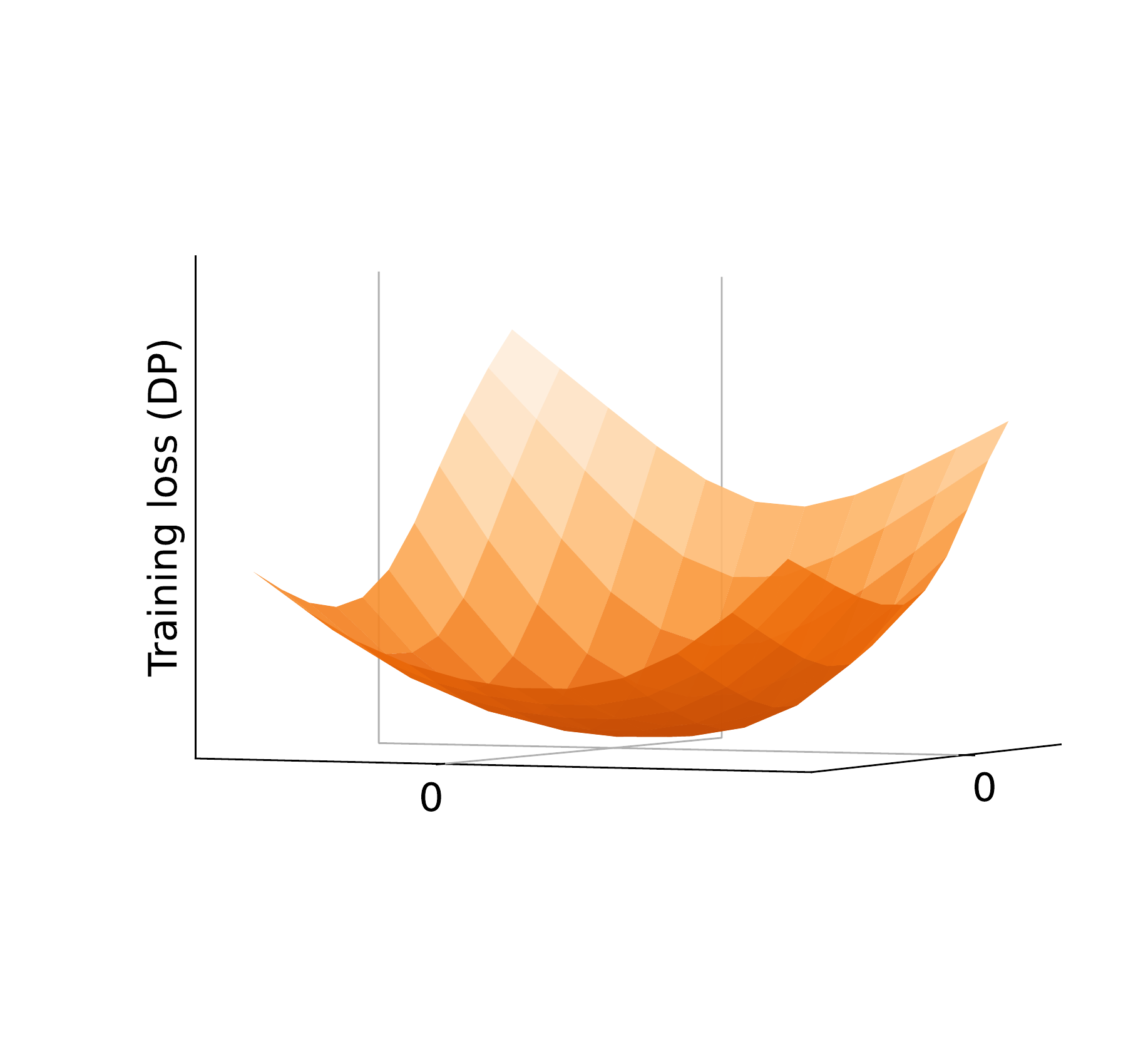}}
    \subfigure[DP-SAT]{\label{fig:sharpness_dp_dpsat}\includegraphics[width=40mm]{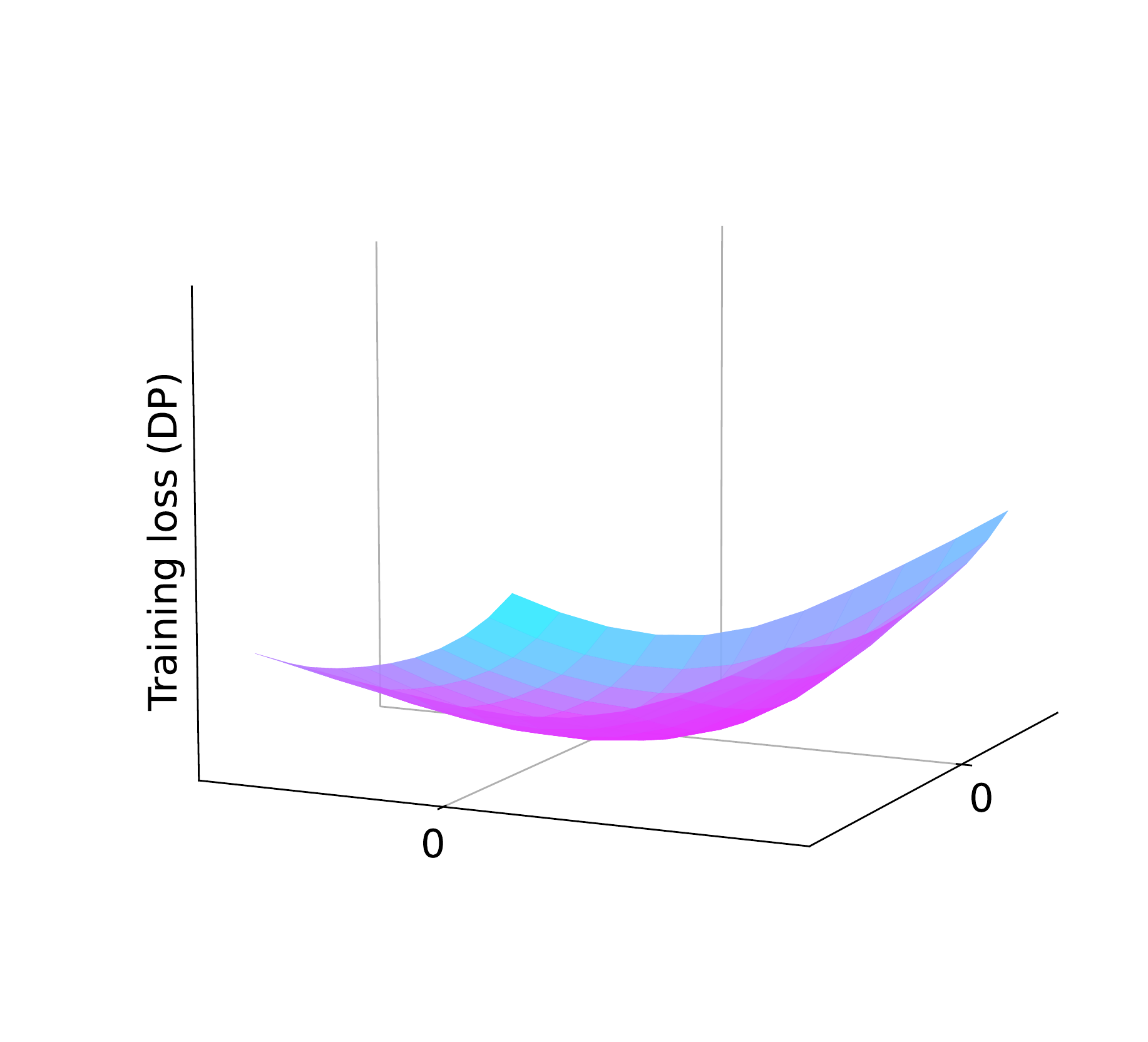}}
    \subfigure[All]{\label{fig:sharpness_dp_all}\includegraphics[width=40mm]{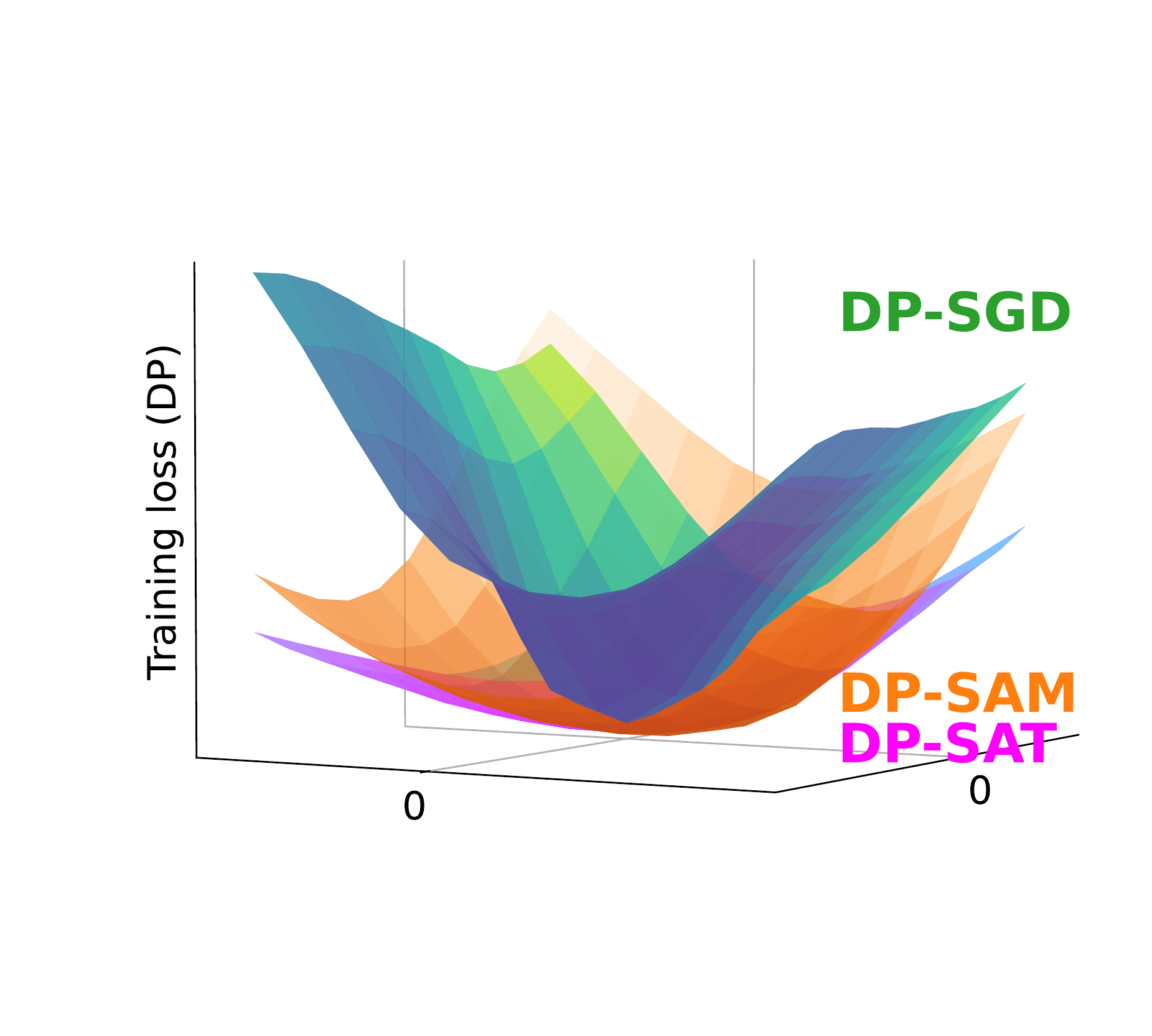}}
    \caption{Visualization of the loss landscapes of DP-SGD, DP-SAM, and DP-SAT.}
    \label{fig:sharpness_dp_plots}
\end{figure*}

\section{Why Ascent Step should be privatized in DP-SAM?}
\label{app:dpsam_private}
For clear understanding, here we denote the non-private parameter as $\boldsymbol {\bar \theta}$ with bar and private parameter $\boldsymbol {\hat \theta}$ with a hat. To satisfy the condition of Moments accountant for DP-SGD, each of the previous weights should be private as $\hat {\boldsymbol \theta}_{t-1}$.  In each step, the differential privacy of  $\boldsymbol \theta_t$ is guranteed by a differentially private mechanism $\mathcal{M}$ , i.e., $\hat{\boldsymbol {\theta}}_{t}=\mathcal M(\hat {\boldsymbol {\theta}}_{t-1};\boldsymbol x)$, where $\boldsymbol x$ is training data samples.

Consider the simple two-step mechanism $\mathcal{M} (\boldsymbol {\hat\theta} ;\boldsymbol x)=  \mathcal{M}_2(\mathcal{M}_1(\boldsymbol {\hat\theta};\boldsymbol x),\boldsymbol {\hat\theta};\boldsymbol  x)$. We argue that $\mathcal{M} (\boldsymbol {\hat \theta};\boldsymbol x)$ is differentially private when $\mathcal{M}_2, \mathcal{M}_1$ are private by the composition theorem w.r.t. training data $\boldsymbol x$.

By contradiction, let only $\mathcal M_2$  is private w.r.t $\boldsymbol x$, $\mathcal{M} (\boldsymbol {\hat\theta} ;\boldsymbol x)=  \mathcal{M}_2(\bar {\boldsymbol \theta}',\boldsymbol {\hat\theta};\boldsymbol  x)$, where the output of $\mathcal{M}_1$ is not private $\bar {\boldsymbol \theta}'$ . Thus, we cannot use the post-processing or calculate the sensitivity of $\mathcal{M}_2$ because $\bar {\boldsymbol \theta}'$ possesses information of $\boldsymbol x$. This means that adding noise only to $\mathcal M_2$ cannot guarantee the complete privacy level. Thus, we should consider the privacy budget consumed by $\mathcal{M}_1$ w.r.t. $\boldsymbol x$.

\section{Why Previous Gradient can be used in DP-SAT?}
\label{app:sat_motivation}

This section demonstrates that DP-SAT can take advantage of sharpness-aware training even when the ascent direction is derived from different batch samples.
The primary motivation is that gradient-based optimization occurs in a low-dimensional subspace of top eigenvalues \cite{sagun2018empirical,papyan2019measurements}. According to \Figref{fig:hessian}, we can see that the Eigenspectrum of DP is separable into two divisions, analogous to non-DP training. The principal subspace consists of a number of classes outliers (about 10 in CIFAR-10) with large eigenvalues, separated from a continuous bulk centered on zero. Exploiting the principal subspace, it is more advantageous to utilize the previous perturbed gradients $\vg_1,\ldots,\vg_{t-1}$ as an ascent direction than random directions. Note that \citet{jang2022reparametrization} and \citet{lee2022implicit} have proposed methods to implicitly regularize the sharpness through the use of Hessian approximation, yielding experimental results similar to SAM. 

\begin{figure}[h]
\centering     
    \subfigure[Impact of clipping value $C$]{\label{fig:cos_sim_C}\includegraphics[width=45mm]{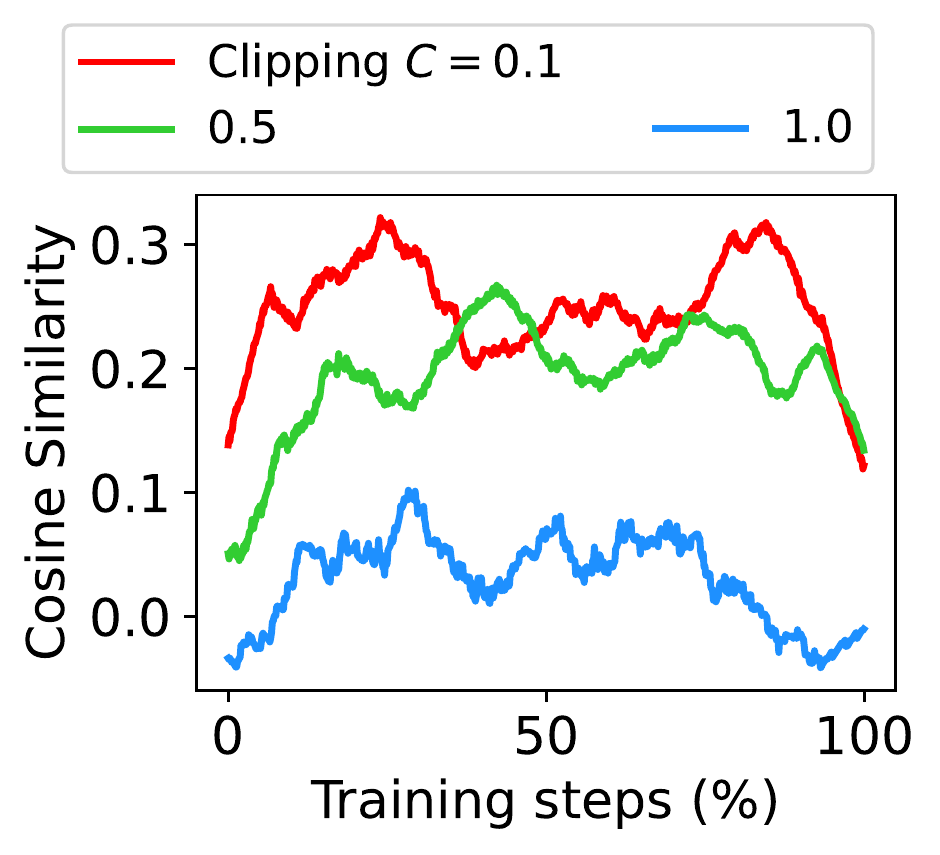}}
    \subfigure[Impact of batch size]{\label{fig:cos_sim_BS}\includegraphics[width=45mm]{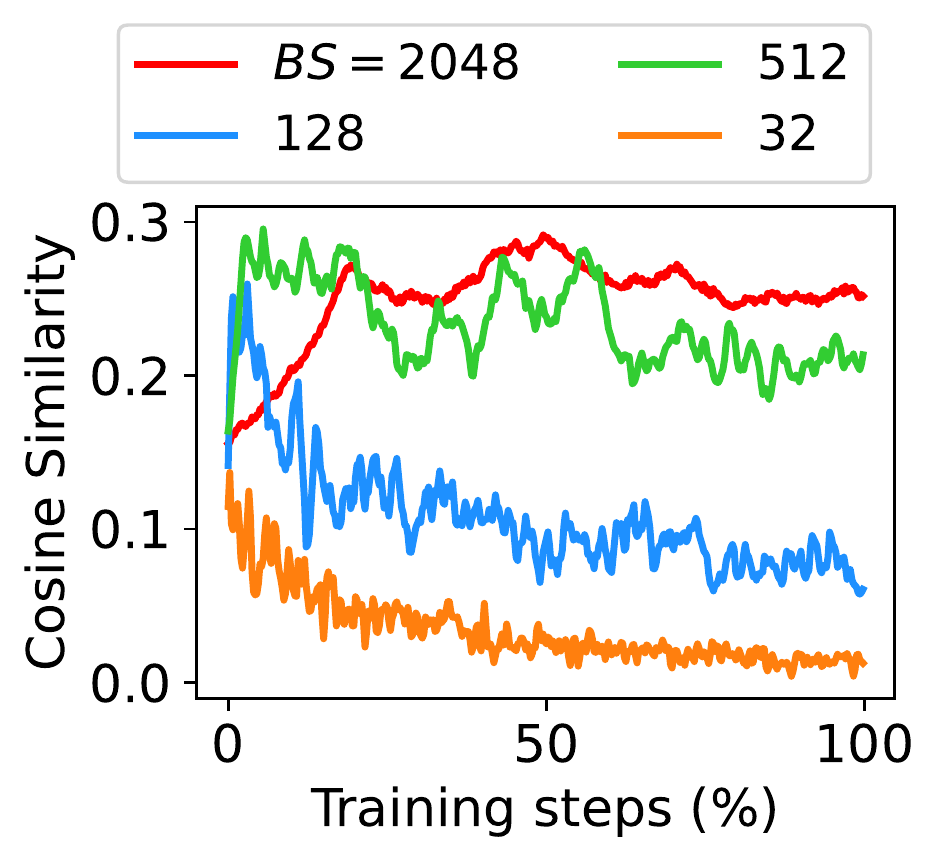}}
    \caption{Cosine similarity of the current ascent direction $\bar{\vdelta}_t$ and previous ascent direction $\bar{\vdelta}_{t-1}$ before adding noise. Small clipping values $C$ and large batch sizes lead to similar gradient directions. The curves are smoothed for better visualization and the x-axis is denoted in percentage because training with smaller batch size has more training steps.}
    \label{fig:cosines}
\end{figure}

The utilization of clipping and large batch size, two properties of DP training, cause the gradients from different batch samples to be alike. 
The clipping operation affects the training dynamics as mentioned in \Secref{sec:dp_training} and DP training usually uses larger batch sizes to decrease the level of injected noise $\sigma$ \cite{tramer2021differentially}. To check the cosine similarities, we now investigate how the clipping value $C$ and batch sizes affect cosine similarities of ascent steps $\bar\vdelta_t=\sum_{i\in I_t} \text{clip}(\nabla \ell_i(\vw_t), C)$, before adding noise. The cosine similarities of current ascent direction $\bar\vdelta_t$ and previous ascent direction $\bar\vdelta_{t-1}$ for the total training procedure are illustrated in \Figref{fig:cosines}. We fixed the batch size of 2048 for experiments on $C$ and fixed $C=0.1$ for experiments on batch size. In general, the larger batch size and the smaller clipping value result in higher cosine similarities of $\bar{\vdelta_t}$ and $\bar\vdelta_{t-1}$. Furthermore, all the cosine similarities drastically rise at the early epochs, which are coherent periods for the sudden drop of train loss in \Figref{fig:training_loss_error}.

We then take the average cosine similarities of  \Figref{fig:cosines} in \Tabref{tab:cos_sim}. In real experimental settings, batch size of 2048 and $C=0.1$, cosine similarity of 0.26 indicates a high level of alignment, given the high dimensionality of deep learning models. These results are in contrast to zero alignments in the case of $C=\infty$ or a batch size of 32. Note that \citet{andriushchenko2022towards} revealed that accurately solving the internal maximization problem has analogous effects to enlarging the radius $\rho$ due to the nonlinearity of the weights space. This might be a reason that the ascent direction can be selected without computing the ascent direction within the same mini-batch.




\begin{table}[h]
\centering
\caption{Average cosine similarities of ascent steps at the current and previous steps during all training steps in \Figref{fig:cosines}.}
\label{tab:cos_sim}
\resizebox{0.82\textwidth/2}{!}{%
\begin{tabular}{c|cccc}
\toprule
C          & $\infty$ & 1  & 0.5  & 0.1   \\ \hline
$\mathbb{E}_t [\cos(\bar{\vdelta}_t,\bar{\vdelta}_{t-1}$)]  & {-} &  \multicolumn{1}{r}{0.03} &\multicolumn{1}{r}{0.20} & \multicolumn{1}{r}{\textbf{0.26}}    \\ \hline\hline
Batch size & 32 & 128 & 512 &   2048     \\ \hline
$\mathbb{E}_t [\cos(\bar{\vdelta}_t,\bar{\vdelta}_{t-1}$)] & \multicolumn{1}{r}{0.03} & \multicolumn{1}{r}{0.10} &  \multicolumn{1}{r}{0.23} & \multicolumn{1}{r}{\textbf{0.26}} 
\\ \bottomrule 
\end{tabular}
}
\end{table}

\section{Ablation study}
\label{app:ablation}
\subsection{Momentum variants}
\label{app:momentum}
Applying the concept of momentum in SAM is widely used to enhance the performance of SAM \cite{du2022sharpnessaware, qu2022generalized}. We also suggest DP-SAT-Momentum, which uses the idea of momentum to calculate the ascent direction in DP-SAT. 
First of all, the momentum updates at $t$ step as follows:
\begin{align}
    \vv_t \leftarrow \gamma\vv_{t-1}+\vg_{t}.
\end{align}
Then, the accumulated momentum can be written as:
\begin{align}
    \vv_t&=\sum_{\tau=1}^{t-1}\gamma^\tau\vg_{t-\tau}.
\end{align}
Then, we can set the ascent direction using all the perturbed gradient information from step $1,\ldots,t-1$ as follows:
\begin{align}
    \vdelta_t^{Momentum}= \rho \frac{\vv_t}{\|\vv_t\|} = \rho\frac{\gamma\vg_{t-1} + \gamma^2 \vg_{t-2} + \cdots +  \gamma^{t-1} \vg_{1}}{\|\gamma\vg_{t-1} + \gamma^{2} \vg_{t-2} + \cdots +  \gamma^{t-1} \vg_{1}\|}.
\end{align}
As we use base optimizers with momentum (which will be discussed in the next subsection), we set the momentum $\gamma$ the same as the momentum of the base optimizer for convenience.
This method is also free from additional privacy costs because all the previous perturbed gradients are guaranteed by post-processing. 

The experimental results of DP-SAT-Momentum are in \Tabref{tab:momentum}. Empirically, we should use the radius $\rho$ for DP-SAT-Momentum as 10 times larger than that of DP-SAT. Both optimization methods show similar experimental effects. 
We believe this phenomenon is that the gradient similarities induced in DP training aforementioned in \Appref{app:sat_motivation} are sufficient to approximate the ascent step, different from non-DP training \cite{du2022sharpnessaware, qu2022generalized}.

\begin{table}[h]
\centering
\caption{Performance of DP-SAT and DP-SAT-Momentum on MNIST, FashionMNIST, and CIFAR-10.}
\label{tab:momentum}
\resizebox{0.75\textwidth}{!}{%
\begin{tabular}{cccrr}
\toprule
\multirow{2}{*}{Datasets} & \multirow{2}{*}{Model} & \multirow{2}{*}{\begin{tabular}[c]{@{}c@{}}Privacy budget $\varepsilon$\\ ($\delta=10^{-5}$)\end{tabular}} & \multicolumn{2}{c}{Optimizers} \\ \cline{4-5} 
 &  &  & \multicolumn{1}{c}{DP-SAT} & \multicolumn{1}{c}{DP-SAT-Momentum} \\ \hline
\multirow{3}{*}{MNIST} & \multirow{3}{*}{\begin{tabular}[c]{@{}c@{}}DPNAS-MNIST\\ (\#Params: 0.21M)\end{tabular}} & $\varepsilon=1$ & 97.96±0.08 & 98.00±0.13 \\
 &  & $\varepsilon=2$ & 98.71±0.09 & 98.83±0.11 \\
 &  & $\varepsilon=3$ & 98.93±0.02 & 98.91±0.11 \\ \hline
\multirow{3}{*}{FashionMNIST} & \multirow{3}{*}{\begin{tabular}[c]{@{}c@{}}DPNAS-MNIST\\ (\#Params: 0.21M)\end{tabular}} & $\varepsilon=1$ & 85.92±0.35 & 85.47±0.42 \\
 &  & $\varepsilon=2$ & 87.75±0.24 & 87.55±0.18 \\
 &  & $\varepsilon=3$ & 88.60±0.04 & 88.56±0.30 \\ \hline
\multirow{3}{*}{CIFAR-10} & \multirow{3}{*}{\begin{tabular}[c]{@{}c@{}}DPNAS-CIFAR10\\ (\#Params: 0.53M)\end{tabular}} & $\varepsilon=1$ & 60.13±0.34 & 60.10±0.46 \\
 &  & $\varepsilon=2$ & 67.23±0.12 & 67.24±0.21 \\
 &  & $\varepsilon=3$ & 69.86±0.49 & 69.63±0.10
\\ \bottomrule 
\end{tabular}
}
\end{table}

\subsection{Without tuning the radius $\rho$}
\label{app:tuningrho}
We provide a baseline for experiments that do not involve tuning the radius $\rho$ in \Tabref{tab:samerho}, particularly with regard to \Tabref{tab:main}. To address this, we set $\rho=0.03$ for the MNIST and FashionMNIST datasets and $\rho=0.01$ for the DPNAS-CIFAR10 in all experiments presented in \Tabref{tab:main}. The other settings except for the radius $\rho$ are the same.
\begin{table*}[ht!]
\centering
\caption{Classification accuracy of DP-SGD, and DP-SAT on the MNIST, FashionMNIST, and  CIFAR-10 datasets with fixing $\rho=0.03$ (except for DPNAS-CIFAR10 architecture with $\rho=0.01$). We bold the highest average accuracy.} 
\label{tab:samerho}
\resizebox{0.6\textwidth}{!}{%
\begin{tabular}{ccccrr}
\toprule
Datasets & Model & $\varepsilon$ & $\rho$ & \multicolumn{1}{c}{DP-SGD} & \multicolumn{1}{c}{DP-SAT} \\ \hline
\multirow{6}{*}{MNIST} & \multirow{3}{*}{GNResnet-10} & 1 & 0.03 & 95.15±0.17 & \textbf{96.00±0.21} \\
 &  & 2 & 0.03 & 96.68±0.27 & \textbf{97.35±0.14} \\
 &  & 3 & 0.03 & 97.30±0.14 & \textbf{97.83±0.10} \\ \cline{2-6} 
 & \multirow{3}{*}{DPNAS-MNIST} & 1 & 0.03 & 97.77±0.13 & \textbf{97.96±0.08} \\
 &  & 2 & 0.03 & 98.60±0.06 & \textbf{98.71±0.09} \\
 &  & 3 & 0.03 & 98.70±0.12 & \textbf{98.93±0.02} \\ \hline
\multirow{6}{*}{FashionMNIST} & \multirow{3}{*}{GNResnet-10} & 1 & 0.03 & 80.57±0.25 & \textbf{81.33±0.45} \\
 &  & 2 & 0.03 & 82.71±0.35 & \textbf{84.53±0.41} \\
 &  & 3 & 0.03 & 84.55±0.17 & \textbf{85.91±0.22} \\ \cline{2-6} 
 & \multirow{3}{*}{DPNAS-MNIST} & 1 & 0.03 & 84.62±0.19 & \textbf{85.52±0.28} \\
 &  & 2 & 0.03 & 86.99±0.57 & \textbf{87.72±0.28} \\
 &  & 3 & 0.03 & 87.97±0.17 & \textbf{88.38±0.23} \\ \hline
\multirow{6}{*}{CIFAR-10} & \multirow{3}{*}{CNN-Tanh with SELU} & 1 & 0.03 & \textbf{45.24±0.42} & \textbf{45.45±0.36} \\
 &  & 2 & 0.03 & \textbf{56.90±0.33} & \textbf{57.34±0.59} \\
 &  & 3 & 0.03 & 61.84±0.48 & \textbf{62.64±0.45} \\ \cline{2-6} 
 & \multirow{3}{*}{DPNAS-CIFAR10} & 1 & 0.01 & 59.42±0.38 & \textbf{60.13±0.34} \\
 &  & 2 & 0.01 & 66.30±0.27 & \textbf{67.23±0.12} \\
 &  & 3 & 0.01 & 68.43±0.43 & \textbf{69.86±0.49}  \\\hline \multirow{3}{*}{SVHN} & \multirow{3}{*}{DPNAS-CIFAR10} & 1 & 0.03 & 82.25±0.15 & \textbf{82.95±0.28} \\
 &  & 2 & 0.03 & 86.85±0.33 & \textbf{87.49±0.15} \\
 &  & 3 & 0.03 & 88.18±0.23 & \textbf{88.74±0.18}                              
\\ \bottomrule
\end{tabular}%
}
\end{table*}

\subsection{Different base optimizers}
The experimental results of using other optimizers, such as SGD without momentum and Adam are in \Tabref{tab:adam}. For Adam, we select the learning rate of 0.0002. The difference is marginal between SGD with a momentum of 0.9 and Adam, showing better performance than SGD without momentum. In all cases, DP-SAT can achieve better performance regardless of base optimizers.
\begin{table*}[ht!]
\centering
\caption{Classification accuracies for DP-SGD, DP-Adam, and DP-SAT on the MNIST, FashionMNIST, and CIFAR-10 datasets. $\beta$ indicates momentum. The bold indicates results within the standard deviation of the highest mean score.}
\label{tab:adam}
\resizebox{0.8\textwidth}{!}{%
\begin{tabular}{ccrrrrr}
\toprule
\multirow{2}{*}{Datasets} &
  \multirow{2}{*}{\begin{tabular}[c]{@{}c@{}}Privacy budget $\varepsilon$\\ ($\delta=10^{-5}$)\end{tabular}} &
  \multicolumn{5}{c}{Optimizer} \\ \cline{3-7} 
 &
   &
  \multicolumn{1}{c}{\begin{tabular}[c]{@{}c@{}}DP-SGD\\ ($\beta=0.0$)\end{tabular}} &
  \multicolumn{1}{c}{\begin{tabular}[c]{@{}c@{}}DP-SGD\\ ($\beta=0.9$)\end{tabular}} &
  \multicolumn{1}{c}{DP-Adam} &
  \multicolumn{1}{c}{\begin{tabular}[c]{@{}c@{}}DP-SAT \\ (SGD, $\beta=0.9$)\end{tabular}} &
  \multicolumn{1}{c}{\begin{tabular}[c]{@{}c@{}}DP-SAT \\ (Adam)\end{tabular}} \\ \hline
\multirow{3}{*}{MNIST} &
  $\varepsilon=1$ &
  97.62±0.13 &
  97.77±0.13 &
  97.96±0.15 &
  97.96±0.08 &
  \textbf{98.30±0.12} \\
 &
  $\varepsilon=2$ &
  97.65±0.22 &
  98.60±0.06 &
  98.54±0.10 &
  \textbf{98.71±0.09} &
  \textbf{98.65±0.10} \\
 &
  $\varepsilon=3$ &
  97.70±0.23 &
  98.70±0.12 &
  98.68±0.07 &
  \textbf{98.93±0.02} &
  98.85±0.09 \\ \hline
\multirow{3}{*}{FashionMNIST} &
  $\varepsilon=1$ &
  80.80±0.24 &
  84.62±0.19 &
  84.85±0.34 &
  \textbf{85.92±0.35 }&
  85.27±0.34 \\
 &
  $\varepsilon=2$ &
  81.55±0.52 &
  86.99±0.57 &
  87.17±0.21 &
  \textbf{87.75±0.24} &
  87.26±0.28 \\
 &
  $\varepsilon=3$ &
  82.03±0.19 &
  87.97±0.17 &
  87.84±0.38 &
  \textbf{88.60±0.04 }&
  88.26±0.40 \\ \hline
\multirow{3}{*}{CIFAR-10} &
  $\varepsilon=1$ &
  56.15±0.74 &
  59.42±0.38 &
  61.75±0.60 &
  60.13±0.34 &
  \textbf{62.52±0.61} \\
 &
  $\varepsilon=2$ &
  56.59±0.79 &
  66.30±0.27 &
  66.68±0.39 &
  \textbf{67.23±0.12} &
  \textbf{67.26±0.36} \\
 &
  $\varepsilon=3$ &
  56.87±0.73 &
  68.43±0.43 &
  68.86±0.39 &
  \textbf{69.86±0.49} &
  69.48±0.38
\\ \bottomrule
\end{tabular}%
}
\end{table*}
\subsection{Effect of radius $\rho$}
\begin{figure*}[!ht]
\centering     
    \includegraphics[width=140mm]{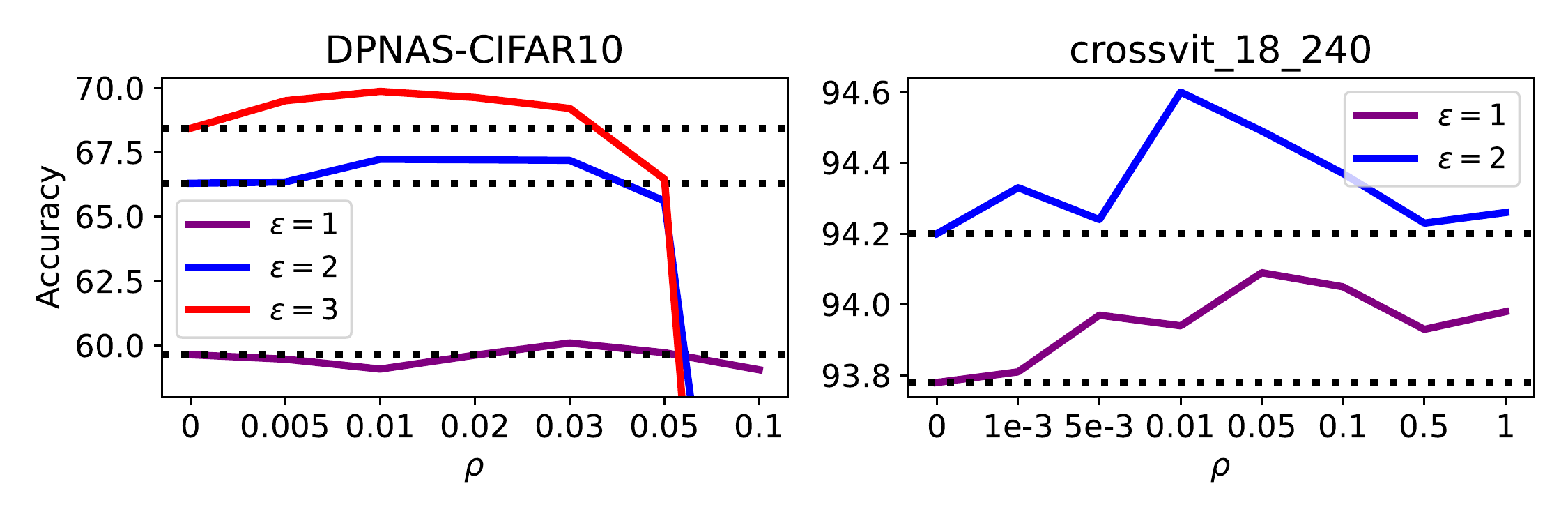}
    \caption{Sensitivity analysis of $\rho$ in DP-SAT on CIFAR-10 dataset. }
    \label{fig:sensitivity_rho}
\end{figure*}

We compare the performance of DP-SAT under different $\rho$. We illustrate the results of various $\rho$ on CIFAR-10 in \Figref{fig:sensitivity_rho}. It shows the importance of finding an appropriate radius in parameter space, where too small or big radius cannot improve the performance. Furthermore, because of the instability of DP training, the accuracy has a high variance and shows some fluctuation in the tendency.

\subsection{Performance of $(2\varepsilon,2\delta)$ DP-SAM}

\Tabref{tab:cifar10_dpsam} shows the performance of DP-SGD and DP-SAM with $\varepsilon=\{1,2,3\}$ of DPNAS-CIFAR models on CIFAR-10. The experimental settings are the same in \Secref{sec:exp}.
DP-SAM shows worse accuracy than DP-SGD due to its extensive privacy consumption on each iteration. 
If we consider $(2\varepsilon,2\delta)$ DP-SAM $^\dagger$, which can guarantee the same training steps $T$ as DP-SGD, then sharpness-aware training achieves better performance in DP training. This indicates that sharpness-aware training can improve performance if we can choose the direction of the ascent step without privacy consumption.

\begin{table}[h]
\centering
\caption{Performance of DP-SGD and DP-SAM on CIFAR-10. DP-SAM shows an accuracy drop because of the doubled privacy budget. The last column $(2\varepsilon,2\delta)$ DP-SAM$^\dagger$ indicates the possible improvements of sharpness-aware training.}
\label{tab:cifar10_dpsam}
\resizebox{0.55\textwidth}{!}{%
\begin{tabular}{c|rrr}
\toprule
\multirow{2}{*}{\begin{tabular}[c]{@{}c@{}}Privacy budget $\varepsilon$\\ ($\delta=10^{-5}$)\end{tabular}} & \multicolumn{3}{c}{Methods} \\ \cline{2-4} 
                & \multicolumn{1}{c}{DP-SGD} & \multicolumn{1}{c|}{DP-SAM}     & \multicolumn{1}{c}{$(2\varepsilon,2\delta)$ DP-SAM $^\dagger$} \\ \hline
$\varepsilon=1$ & 59.42±0.38                 & \multicolumn{1}{r|}{45.15±0.63} & 60.18±0.52                     \\
$\varepsilon=2$ & 66.30±0.27                & \multicolumn{1}{r|}{58.85±0.70} & 66.74±0.37                     \\
$\varepsilon=3$ & 68.43±0.43                & \multicolumn{1}{r|}{63.59±0.25} & 69.58±0.27  \\ \hline\hline
Training epochs & 30 & \multicolumn{1}{r|}{7.07} & 30
\\ \bottomrule 
\end{tabular}
}
\end{table}

\subsection{Convergence analysis}
\begin{figure*}[h]
\centering     
    \includegraphics[width=120mm]{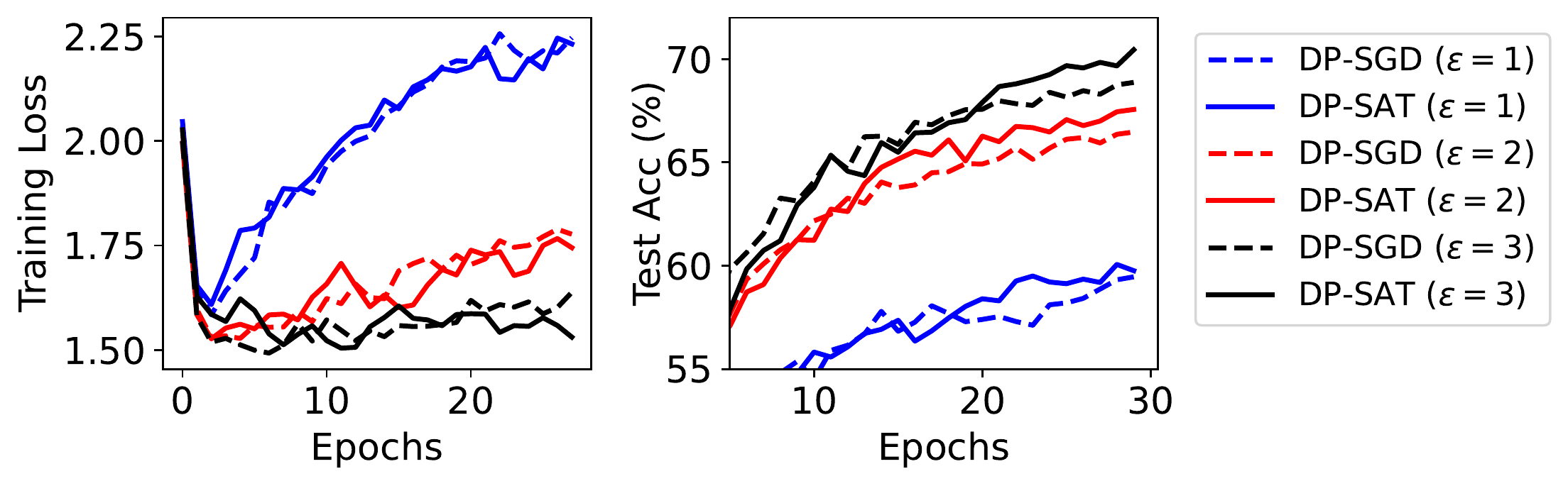}
    \caption{Training loss and Test accuracy of DP-SGD and DP-SAT by varying $\varepsilon$ on CIFAR-10.}
    \label{fig:training_loss_error_cifar}
\end{figure*}

\begin{figure*}[h]
\centering     
    \includegraphics[width=120mm]{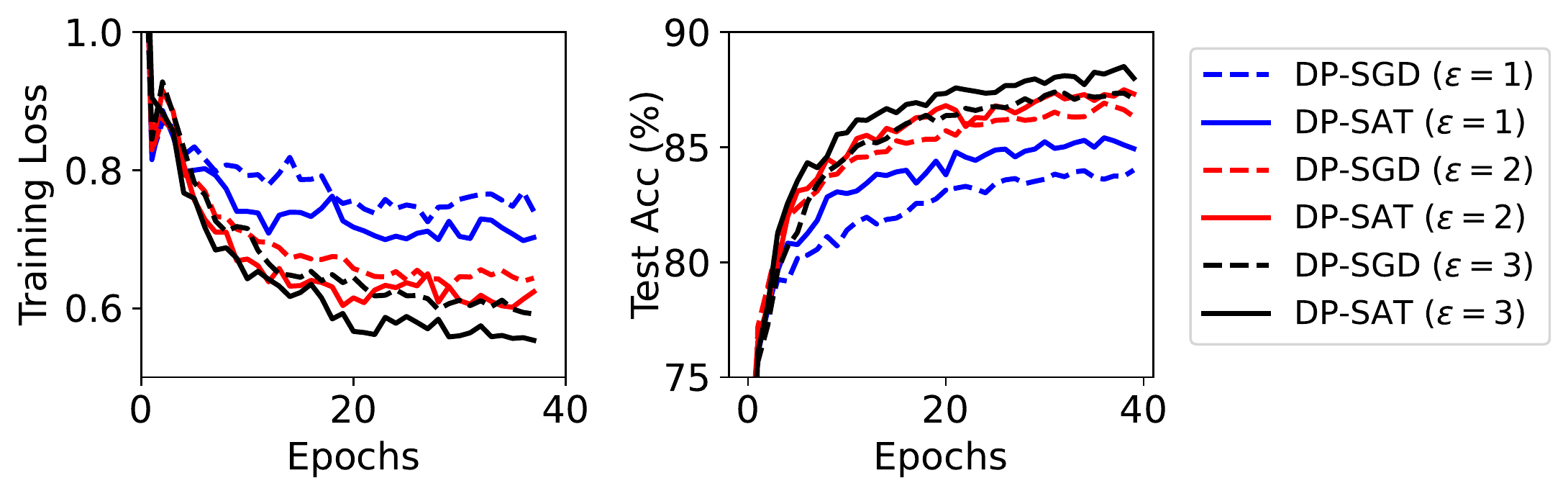}
    \caption{Training loss and Test accuracy of DP-SGD and DP-SAT by varying $\varepsilon$ on MNIST.}
    \label{fig:training_loss_error_mnist}
\end{figure*}

We illustrate the convergence of training loss and corresponding test accuracy of DP-SGD and DP-SAT on CIFAR-10 and MNIST in Figures \ref{fig:training_loss_error_cifar} and \ref{fig:training_loss_error_mnist}. The convergence speed is a little bit slower than DP-SGD but it can reach lower training loss, which is a similar phenomenon of SGD and SAM as depicted in \cite{kaddour2022flat}. 

\subsection{Accuracy plot of \Figref{fig:acc_dpnondp}}
\begin{figure*}[h]
\centering     
    \includegraphics[width=160mm]{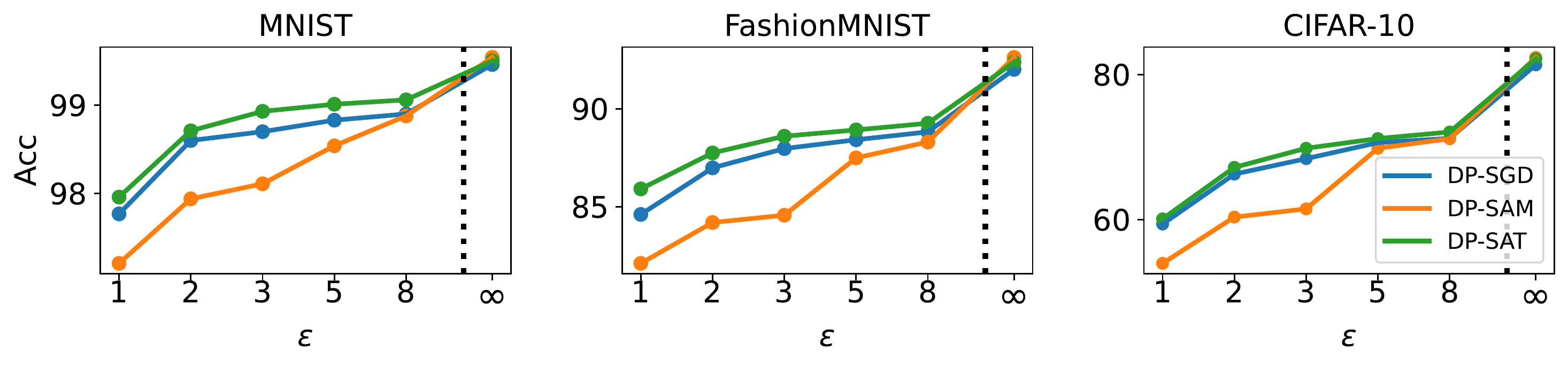}
    \caption{Accuracy for DP-SAM and DP-SAT w.r.t DP-SGD. $\varepsilon=\infty$ indicates non-DP settings.}
    \label{fig:acc_dpnondp_acc}
\end{figure*}

We illustrate the accuracy plot of  \Figref{fig:acc_dpnondp} in \Figref{fig:acc_dpnondp_acc}.

\end{document}